\documentclass{article}
\usepackage[margin=1in]{geometry}
\usepackage{microtype}
\usepackage{mathpazo}
\usepackage[T1]{fontenc}

\usepackage{amsmath,amsfonts,bm}

\def\eqref#1{equation~\ref{#1}}

\def\1{\bm{1}}

\def\vtheta{{\bm{\theta}}}

\def\vg{{\bm{g}}}
\def\vh{{\bm{h}}}

\def\mG{{\bm{G}}}
\def\mH{{\bm{H}}}

\DeclareMathAlphabet{\mathsfit}{\encodingdefault}{\sfdefault}{m}{sl}
\SetMathAlphabet{\mathsfit}{bold}{\encodingdefault}{\sfdefault}{bx}{n}

\usepackage{algorithm}
\usepackage{algpseudocode}
\usepackage{graphicx}
\usepackage[table,dvipsnames]{xcolor}
\definecolor{stanfordCardinal}{HTML}{8C1515}
\usepackage{booktabs}
\usepackage{multirow}
\usepackage{listings}
\usepackage{caption}
\usepackage{enumitem}
\usepackage[toc,page,header]{appendix}
\usepackage{titletoc}
\usepackage{url}
\usepackage{xspace}
\usepackage{tikz}
\usepackage{fontawesome5}
\usepackage[pdfusetitle,linktocpage,colorlinks=true,linkcolor=stanfordCardinal,citecolor=stanfordCardinal,urlcolor=stanfordCardinal]{hyperref}
\usepackage[backend=biber,style=alphabetic,natbib=true,maxbibnames=99,minalphanames=3]{biblatex}
\definecolor{methodSync}{HTML}{BCD1CA}
\definecolor{methodAsync}{HTML}{CBCADB}
\definecolor{methodFullyAsync}{HTML}{EBCECE}
\definecolor{methodThunderSync}{HTML}{CC785C}
\newcommand{\methoddot}[1]{\textcolor{#1}{\raisebox{0.3pt}{\setlength{\unitlength}{1pt}\begin{picture}(5,5)\put(2.5,2.5){\circle*{5}}\end{picture}}}\hspace{0.35em}}

\setlist[itemize]{leftmargin=*,labelindent=1em}
\setlist[enumerate]{leftmargin=*,labelindent=1em}

\makeatletter
\renewcommand{\paragraph}{%
  \@startsection{paragraph}{4}{\z@}%
  {1.5ex \@plus 1ex \@minus .2ex}{-1em}%
  {\normalfont\normalsize\bfseries}%
}
\makeatother

\newcommand{\sysnameraw}{ThunderSync}
\newcommand{\sysname}{\sysnameraw\xspace}
\newcommand{\arxivversion}{}
\title{\sysnameraw: Lossless Acceleration\texorpdfstring{\\}{ }of~Agentic Reinforcement Learning}
\DeclareRobustCommand{\authordagger}{%
  {\fontfamily{ptm}\selectfont\textdagger}%
}
\DeclareRobustCommand{\authordoubledagger}{%
  {\fontfamily{ptm}\selectfont\textdaggerdbl}%
}
\makeatletter
\DeclareRobustCommand{\authornotemark}[1]{%
  \begingroup
    \def\@makefnmark{\hbox{\@textsuperscript{\normalfont\@thefnmark}}}%
    \ifnum#1=2\relax
      \renewcommand\thefootnote{\authordagger}%
    \fi
    \ifnum#1=3\relax
      \renewcommand\thefootnote{\authordoubledagger}%
    \fi
    \footnotemark[#1]%
  \endgroup
}
\makeatother
\author{%
  Seil Kang\textsuperscript{1,2,}\authornotemark{1},\hspace{0.65em}%
  Hangoo Kang\textsuperscript{1,}\authornotemark{1},\hspace{0.65em}%
  Tarun Suresh\textsuperscript{1},\hspace{0.65em}%
  Youngeun Kim\textsuperscript{3} \\
  Shreyas Pimpalgaonkar\textsuperscript{4},\hspace{0.65em}%
  Seong Jae Hwang\textsuperscript{2,}\authornotemark{2},\hspace{0.65em}%
  Azalia Mirhoseini\textsuperscript{1,}\authornotemark{3} \\[6pt]
  {\small \textsuperscript{1}Stanford University \quad \textsuperscript{2}Yonsei University \quad \textsuperscript{3}Korea University \quad \textsuperscript{4}Bespoke Labs} \\[7pt]
\newcommand*{\paperBlogURL}{https://seilk.github.io/thundersync}
\newcommand*{\paperGitHubURL}{https://github.com/seilk/thundersync}
\href{\paperBlogURL}{%
  \begin{tikzpicture}[baseline=(blog.base)]
    \node[inner xsep=4pt, inner ysep=2.5pt, text height=7pt, text depth=2pt, text=stanfordCardinal,
      font=\fontsize{8.7}{10}\selectfont] (blog)
      {\faIcon[regular]{file-alt}\hspace{5pt}{\fontfamily{phv}\bfseries\selectfont Blog post}};
    \draw[stanfordCardinal!35, line width=0.4pt]
      (blog.south west) -- (blog.south east);
    \draw[stanfordCardinal, line width=1.1pt]
      (blog.south west) -- ++(10pt,0);
  \end{tikzpicture}%
}%
\hspace{12pt}%
\href{\paperGitHubURL}{%
  \begin{tikzpicture}[baseline=(github.base)]
    \node[inner xsep=4pt, inner ysep=2.5pt, text height=7pt, text depth=2pt, text=stanfordCardinal,
      font=\fontsize{8.7}{10}\selectfont] (github)
      {\faGithub\hspace{5pt}{\fontfamily{phv}\bfseries\selectfont GitHub}};
    \draw[stanfordCardinal!35, line width=0.4pt]
      (github.south west) -- (github.south east);
    \draw[stanfordCardinal, line width=1.1pt]
      (github.south west) -- ++(10pt,0);
  \end{tikzpicture}%
}

}
\makeatletter
\g@addto@macro\@thanks{%
  \footnotetext[1]{Core contributor.}%
  \begingroup
    \renewcommand\thefootnote{\authordagger}%
    \footnotetext[2]{Co-corresponding author.}%
  \endgroup
  \begingroup
    \renewcommand\thefootnote{\authordoubledagger}%
    \footnotetext[3]{Main corresponding author.}%
  \endgroup
}
\makeatother
\hypersetup{pdfauthor={Seil Kang, Hangoo Kang, Tarun Suresh, Youngeun Kim, Shreyas Pimpalgaonkar, Seong Jae Hwang, Azalia Mirhoseini}}

\date{}

\begin{document}
\maketitle
\begin{abstract}
Language models are moving beyond generating answers to pursuing long-horizon goals in interactive environments. Post-training these agents requires long, heterogeneous trajectories, and synchronous systems leave learner engines idle until rollout and verification finish. To squeeze out these pipeline bubbles, asynchronous training overlaps rollout and learning across updates, but comes at the cost of policy staleness. We introduce \emph{\sysname}, which starts gradient computation as soon as all required inputs are fixed, without policy staleness. For group relative policy optimization (GRPO), \sysname computes each trajectory's score gradient as soon as the reward for that trajectory arrives, without waiting for the group. For on-policy distillation (OPD), it computes gradients for each completed agentic turn's teacher-scored actions while tool calls run in the sandbox. We prove that gradient streaming produces the same GRPO and OPD updates as batch-synchronous training, without changing either objective. On SWE-bench Verified and Terminal Bench 4.0, we train models to the same performance up to $1.9 \times$ faster than synchronous training. With zero policy staleness, \sysname also outperforms asynchronous training at a fixed budget by up to $2.47$ percentage points.
\end{abstract}

\begin{figure}[h!]
  \centering
  \includegraphics[width=1.0\linewidth]{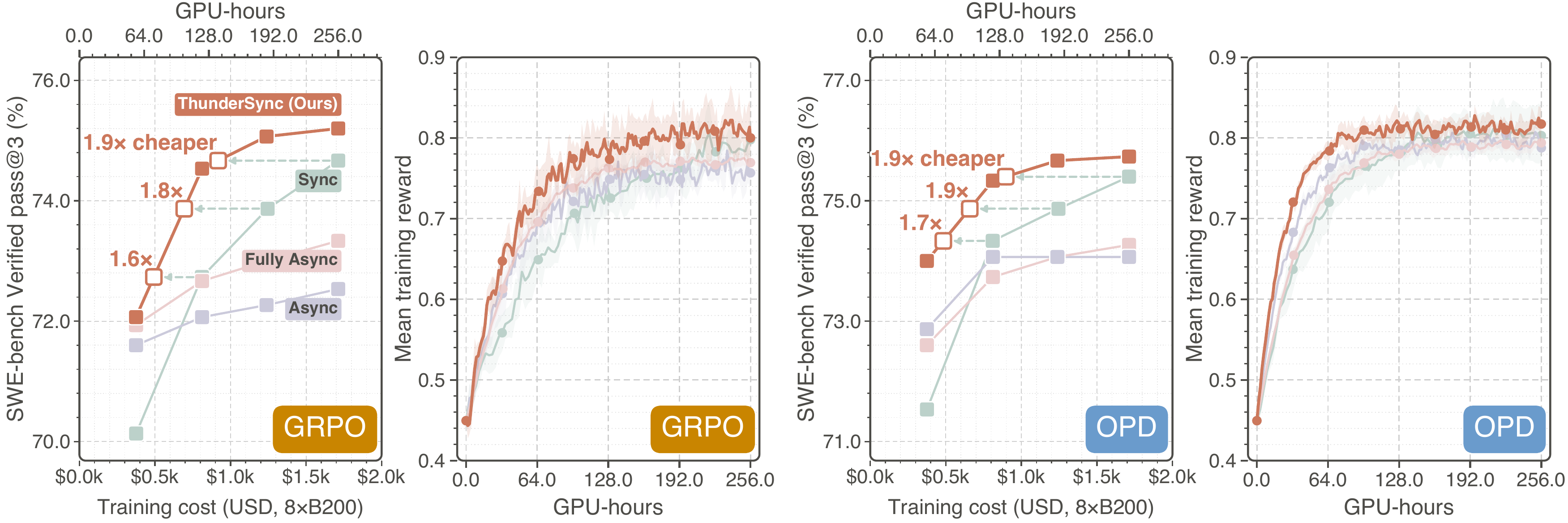}
  \caption{\sysname matches Sync's SWE-bench Verified pass@3 at up to $1.9\times$ lower training cost under GRPO (left pair) and OPD (right pair). At a 128-GPU-hour budget, its zero-staleness schedule improves pass@3 by up to 2.47 percentage points over Async and Fully Async, both of which admit stale rollouts.}
  \label{fig:headline}
\end{figure}

\section{Introduction}
\label{sec:introduction}

Language models are increasingly trained as agents for long-horizon decision-making in interactive environments, reasoning and using tools in response to feedback. Reinforcement learning (RL) from environment or verifier feedback has emerged as a promising way to improve this behavior~\citep{jin2025searchr1,feng2025retool,pan2025swegym,wei2025swerl,gehring2025rlef}. However, training these agents requires interaction trajectories that are expensive to generate and heterogeneous in duration, with long-tailed completion times~\citep{hu2026jigsawrl,chen2026libra,ren2026tiderl}. In batch-synchronous RL, learner engines remain idle until rollout engines finish the batch and all rewards arrive~\citep{fu2025areal,ren2026tiderl}. We found that learner GPUs sit idle for 49.5\% of step time in synchronous GRPO and OPD training while they wait for rollout and verification (Figure~\ref{fig:bubble-motivation}). Prior work reports idle fractions as high as 81\%~\citep{ren2026tiderl}, and the slowest 10\% of rollout requests can take up to half of rollout time~\citep{qin2025seer}, which itself accounts for 70\% to over 90\% of training time~\citep{gao2025rollpacker,zhou2025april}.

Existing systems reduce these pipeline bubbles, idle gaps between dependent stages, by overlapping computation and coordinating resource allocation~\citep{mei2024realhf,sheng2025hybridflow,zhong2025rlhfuse,fu2025areal,han2025asyncflow}. Asynchronous execution overlaps rollout and learning across updates, but some trajectories come from older policies. This policy staleness can require extra algorithmic machinery, including importance-weighted off-policy correction, filtering, clipping, or explicit staleness control~\citep{espeholt2018impala,fu2025areal,han2025asyncflow,hou2026sao}. On-policy pipelines overlap execution within a fixed rollout, but complete-group schedules still delay trajectory-local backward passes until all group rewards arrive~\citep{chen2026rolloutpipe}.

We argue that the apparent trade-off between pipeline overlap and zero staleness is artifactual, arising from coarse execution barriers. Although a synchronous update requires the complete batch, its gradient computation can begin earlier. For example, in group relative policy optimization (GRPO)~\citep{shao2024deepseekmath}, group rewards determine how to weight each trajectory's score gradient, but the gradient can be computed before all rewards arrive. We therefore ask: \emph{can learner computation begin as soon as its objective dependencies are satisfied with zero staleness?}

We introduce \emph{\sysname}, which streams gradients through dependency-aware scheduling of learner work. During update $k$, trajectories are sampled from $\pi_{\theta_k}$ and gradients are evaluated at $\theta_k$. For GRPO, a trajectory's advantage seems to need every reward in its group. \sysname nevertheless starts that trajectory's backward pass as soon as its own reward arrives, and two running sums of score gradients recover the exact group gradient once the group closes (Section~\ref{sec:gradient-streaming}). For on-policy distillation (OPD)~\citep{agarwal2024onpolicydistillation}, \sysname releases a turn's student backward pass once a frozen teacher scores the turn's action tokens, allowing overlap with tool execution and later turns. After all trajectories, groups, and gradient work finish, \sysname combines the gradients, takes one optimizer step, and publishes $\theta_{k+1}$. \sysname is therefore asynchronous in execution and synchronous in policy updates.

We use SWE-smith and SETA-Env for post-training~\citep{yang2025swesmith,shen2026seta} and evaluate on SWE-bench Verified and Terminal Bench 4.0~\citep{jimenez2024swebench,merrill2026terminalbench}. Our scheduling analysis predicts step wall time speedup approaching $2\times$ over synchronous execution when rollout and learner work are balanced and the residual tail and publication cost vanish. With a fixed trajectory batch size, shorter steps let \sysname learn from more trajectories within the same GPU-hours, potentially improving end-to-end performance. Figure~\ref{fig:headline} compares \sysname with Sync by training cost: each arrow marks where \sysname reaches the pass@3 of a Sync checkpoint, at $1.6$ to $1.9\times$ lower cost.

Our main contributions are as follows:

\begin{itemize}
    \item We introduce \sysname and gradient streaming, overlapping rollout and learner computation while preserving a synchronous optimizer step and zero policy staleness.
    \item We derive exact gradient reductions for GRPO and turn-level OPD, allowing early backward computation while preserving their batch-synchronous updates.
    \item We evaluate \sysname on SWE-bench Verified and Terminal Bench 4.0 under GRPO and OPD. \sysname reaches the common pass@3 targets up to 1.9× faster than Sync while retaining zero policy staleness. At 128 GPU-hours, \sysname achieves the highest pass@3 in all settings, 1.27 to 2.47 points above Async.
\end{itemize}

\section{Related Work}
\begin{figure}[t!]
  \centering
  \includegraphics[width=\linewidth]{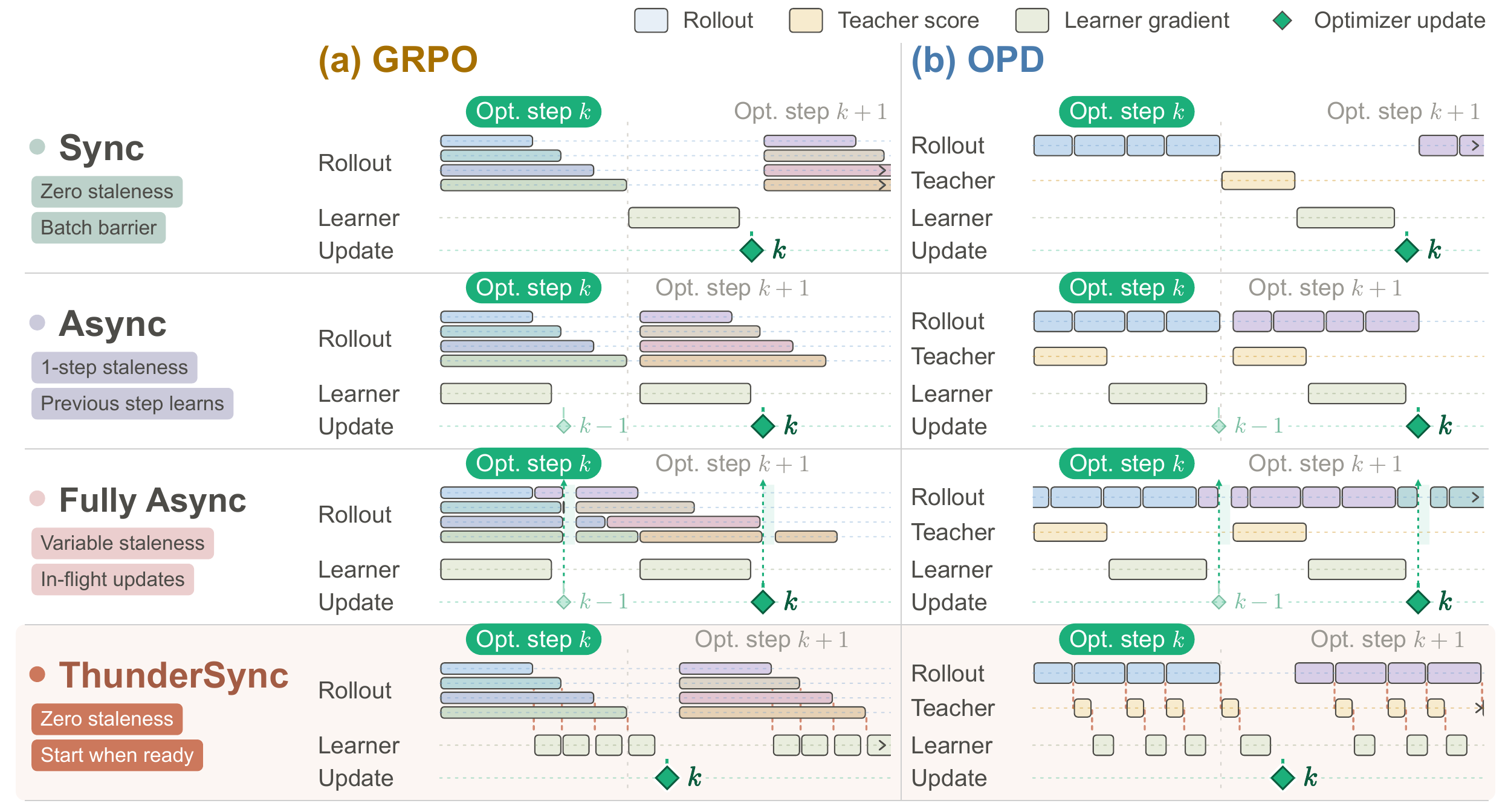}
  \caption{Schematic schedules for GRPO (a) and OPD (b).
  \textcolor[HTML]{BCD1CA}{$\bullet$}~\textbf{Sync} waits for full batches;
  \textcolor[HTML]{CBCADB}{$\bullet$}~\textbf{Async} overlaps adjacent batches
  with one-step staleness;
  \textcolor[HTML]{EBCECE}{$\bullet$}~\textbf{Fully Async} updates rollout weights
  while trajectories remain in flight.
  \textcolor[HTML]{CC785C}{$\bullet$}~\textbf{\sysname} streams ready gradients
  with one batch-synchronous optimizer step and zero staleness.}
  \label{fig:method-overview}
\end{figure}

\subsection{Post-training with Reinforcement Learning Towards Agentic Tasks}
\label{sec:related-agentic-rl}

RL post-training extends single-turn generation to agents that learn through long-horizon interaction with external environments. ArCHer develops hierarchical multi-turn RL, while DigiRL combines offline and online RL for mobile-device control~\citep{zhou2024archer,bai2024digirl}. Web and search agents learn through self-evolving curricula, browser interaction, and outcome rewards~\citep{qi2025webrl,wei2025webagentr1,jin2025searchr1,song2025r1searcher,chen2025research,zheng2025deepresearcher}. Tool-use research studies strategic interpreter use, fine-grained rewards, planning supervision, and dynamic user interaction~\citep{feng2025retool,qian2025toolrl,li2025rltr,zhao2025muarl}. Further work addresses trajectory-level optimization failures, multi-step supervision, and procedural or capability-targeted training environments~\citep{wang2025ragen,goldie2025syntheticdatageneration,ding2026gtpo,sullivan2025randomworld,kang2026tracecapabilitytargetedagentictraining}. Software agents learn from executable environments, software-evolution data, and execution feedback~\citep{pan2025swegym,wei2025swerl,gehring2025rlef}. Agent Lightning and VerlTool provide reusable infrastructure across heterogeneous agent workflows~\citep{luo2025agentlightning,jiang2025verltool}.

\subsection{Optimization of Reinforcement Learning}

Distributed RLHF systems coordinate generation, reference and reward inference, and actor and critic training, using model placement, device reallocation, and stage fusion to reduce idle time~\citep{yao2023deepspeedchat,hu2024openrlhf,mei2024realhf,sheng2025hybridflow,zhong2025rlhfuse}. Building on actor-learner decoupling in distributed RL~\citep{espeholt2018impala,petrenko2020samplefactory}, LlamaRL, AReaL, AsyncFlow, and StreamRL support disaggregated generation and training, streaming execution, asynchronous weight synchronization, and control of policy staleness~\citep{wu2025llamarl,fu2025areal,han2025asyncflow,zhong2025streamrl}. For heterogeneous agentic rollouts, PipelineRL and RolloutPipe pipeline rollout and learner execution~\citep{piche2025pipelinerl,chen2026rolloutpipe}. JigsawRL, Libra, TideRL, and WAR address fine-grained scheduling, elastic resource allocation, long tails, and inference caching~\citep{hu2026jigsawrl,chen2026libra,ren2026tiderl,xu2026war}. SAO performs asynchronous optimization from individually completed rollouts~\citep{hou2026sao}. \sysname schedules learner computation as each objective source becomes ready, overlapping it with unfinished rollouts while preserving a fixed policy snapshot and a single synchronous optimizer step. The agent, reward, and optimization objective remain unchanged.

\section{Methods}
We present \sysname, a gradient streaming framework for zero-stale agentic post-training (Figure~\ref{fig:method-overview}). \sysname releases learner computation when each source becomes \emph{ready}. Exact gradient reduction and a synchronous optimizer barrier preserve the update semantics of GRPO and on-policy distillation (OPD) while allowing rollout and learning to overlap.

\subsection{Background}
\label{sec:background}
\paragraph{Group relative policy optimization (GRPO).} Given a prompt, GRPO samples a group of trajectories and normalizes their rewards within the group to obtain relative advantages. The policy update weights each trajectory's scored action-token log probabilities by its detached advantage.

\paragraph{On-policy distillation (OPD).} OPD generates trajectories with the student policy and evaluates the student's action tokens under a frozen teacher. The resulting token-level supervision defines a distillation loss for the action tokens of each agentic turn.

\begin{figure}[t!]
  \centering
  \includegraphics[width=\linewidth]{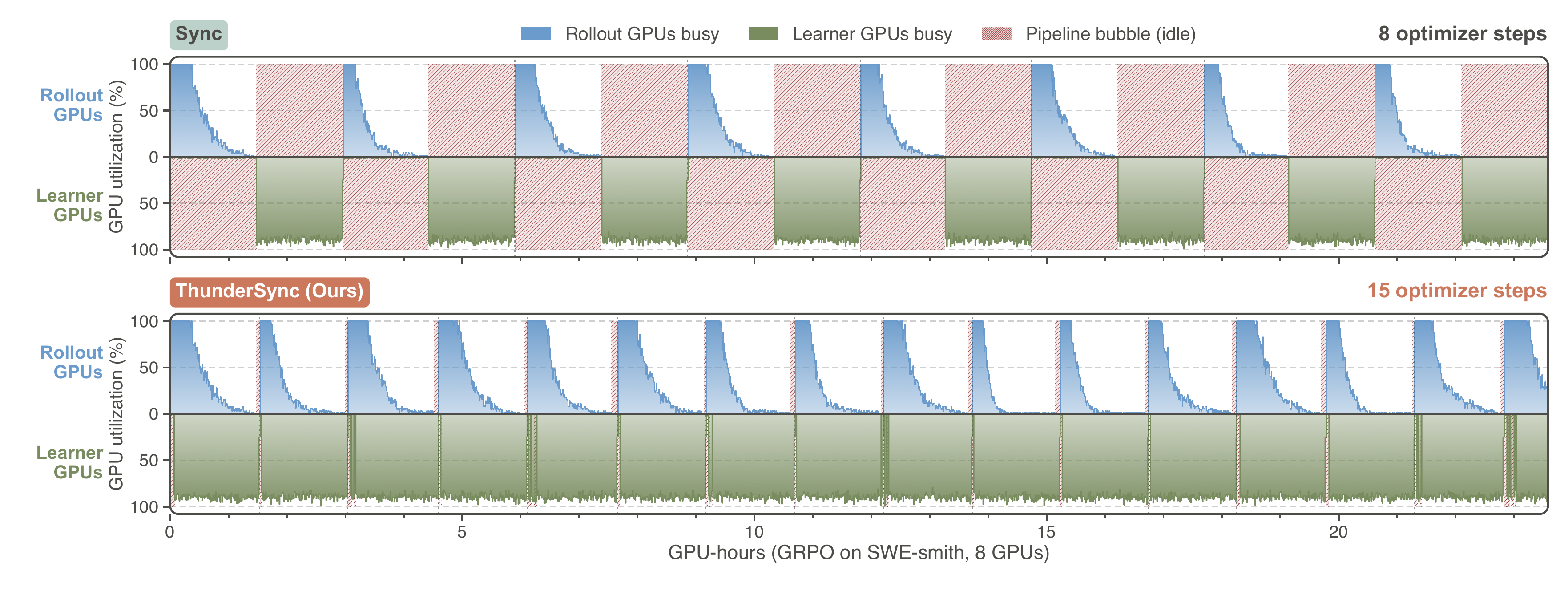}
  \caption{Pipeline bubbles in synchronous agentic RL.
  GPU utilization during GRPO training on SWE-smith, with four rollout GPUs
  above the axis and four learner GPUs below. Rollout utilization (busy share
  of generation slots) drops as long-tail trajectories finish.
  \raisebox{-0.1ex}{\includegraphics[height=1.4ex]{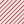}}~Hatched
  bands mark idle time.
  \textcolor[HTML]{BCD1CA}{$\bullet$}~\textbf{Sync} learners wait for the
  last reward for 49.5\% of each step, and rollout GPUs wait while the
  learner trains.
  \textcolor[HTML]{CC785C}{$\bullet$}~\textbf{\sysname} starts each
  trajectory's backward pass when its reward arrives and completes
  15 optimizer steps in the GPU-hours Sync needs for 8.}
  \label{fig:bubble-motivation}
\end{figure}

\paragraph{Actor-learner execution protocols.} Agentic reinforcement learning alternates between two expensive stages: generating long, tool-using trajectories and optimizing the policy from the resulting training signals. Conventional synchronous training places a batch barrier between these stages, leaving the learner idle until every trajectory and reward is available (Figure~\ref{fig:bubble-motivation}). Fully asynchronous methods remove this idle time, but typically update the policy while older rollouts remain in flight, introducing policy staleness. \sysname follows a single principle: each computation is released at the earliest time when its mathematical dependencies are satisfied.

\subsection{Problem Formulation}
\paragraph{Objective sources.} Consider update $k$ with policy $\pi_{\vtheta_k}$ and logical rollout batch $\mathcal{B}_k$. Each trajectory is a sequence of agentic turns, where a turn contains a policy-generated action and the resulting environment or tool observation. We call an independently releasable unit of learner computation an \emph{objective source}. Source $j$ is \emph{ready} at time $\tau_j$ once all its inputs are fixed.

\paragraph{Release semantics.} The ready unit depends on the training objective. For a GRPO group $\mathcal{G}$ of $m$ trajectories, source $i$ is associated with trajectory $i$ and becomes ready when its token sequence, scored-token mask, and reward $r_i$ are fixed. The group statistics remain unavailable until all $m$ rewards arrive. In OPD, the scored action tokens of each turn form a source. Once these tokens and their mask are fixed, the frozen teacher can score them using their realized causal context. The teacher scores release the corresponding student backward pass, which can overlap with tool execution and subsequent turns.

\paragraph{Scheduling objective.} Our problem is to release each source at its earliest objective-valid time while preserving the corresponding batch-synchronous update. All trajectories in update $k$ are sampled from $\pi_{\vtheta_k}$, and all gradients are evaluated at the fixed parameters $\vtheta_k$. After the logical batch closes and all gradient work finishes, the optimizer takes one step to produce $\vtheta_{k+1}$. The logical batch and objective remain unchanged; only the timing of eligible learner work differs.

\subsection{\sysnameraw{} Framework}
\label{sec:framework}

\paragraph{Release rule.}
\sysname releases each source's learner computation at its ready time $\tau_j$, which can precede the completion of its group or of the batch; gradients accumulate at $\vtheta_k$.

\paragraph{Readiness-driven scheduling.}
Ready sources enter a work-conserving queue in readiness order; admission never waits for a smaller trajectory index, or for an unrelated group below a per-rank cap (Appendix~\ref{app:configurations}). Independent groups may be reduced concurrently, while each group preserves a deterministic reduction order. We include deferred gradient propagation at trajectory or group closure as learner work with its own release time. Index learner work items in service order; let $\tau_j$ and $p_j$ denote the release time and service time of item $j$. On a single learner lane with $c_0=0$, completion time satisfies
\begin{equation}
    c_j = \max\{\tau_j,c_{j-1}\}+p_j.
\end{equation}
For $n$ learner work items, let $R_k$ be the time at which the complete rollout batch closes and $P_k$ the common final reduction, optimizer, and publication suffix. The update endpoint is
\begin{equation}
    T_k = \max\{R_k,c_n\}+P_k.
\end{equation}
A batch-synchronous method postpones all learner work until $R_k$. For GRPO, a group-ready pipeline releases all sources in a group at its final reward. \sysname uses each source's ready time for either objective, moving eligible learner work into rollout intervals and reducing the learner tail after the final trajectory.

\paragraph{Zero-stale optimizer barrier.}
\sysname waits for every registered trajectory and group to close and for all deferred gradient propagation to finish. Distributed reduction and objective-specific normalization then produce the batch update direction $\vg_k$ (Section~\ref{sec:gradient-streaming}). \sysname clips this direction and applies exactly one optimizer step,
\begin{equation}
    \vtheta_{k+1}
    =
    \operatorname{Opt}
    \left(
        \vtheta_k,
        \operatorname{clip}(\vg_k)
    \right).
\end{equation}
\sysname publishes the resulting adapter or policy parameters to the rollout replicas. Each replica loads the new version, clears policy-dependent state, and acknowledges that it serves $\vtheta_{k+1}$ before the next update begins. Every rollout and learner source within update $k$ is therefore bound to $\vtheta_k$, and no trajectory is optimized against a policy that changed while it was in flight. Appendix~\ref{app:implementation} gives the full algorithm and the GRPO and OPD implementations.

\subsection{Objective-Specific Gradient Streaming}
\label{sec:gradient-streaming}

\paragraph{Streaming the GRPO gradient.}
The main obstacle to streaming the GRPO gradient is that the advantage of one trajectory depends on the rewards of its entire group. Let $\mathcal{G}$ be a group of $m$ trajectories. For trajectory $i\in\mathcal{G}$, define its masked policy score and corresponding source gradient as
\begin{equation}
    S_i(\vtheta)
    =
    \sum_{t\in\mathcal{M}_i}
    w_{i,t}\log\pi_{\vtheta}
    \!\left(a_{i,t}\mid \vh_{i,t}\right),
    \qquad
    \mH_i
    =
    \left.\nabla_{\vtheta}S_i(\vtheta)\right|_{\vtheta=\vtheta_k},
\end{equation}
where $\mathcal{M}_i$ is the scored action-token set, $w_{i,t}$ the token weighting, and $r_i$ the trajectory reward. We set $w(i,t) = 1/(|B_k| L_i)$, where $L_i$ is the scored-token count of trajectory $i$ and $|B_k|$ is the number of groups in the batch (32 in our experiments), following the GRPO reduction~\cite{shao2024deepseekmath} The reductions differ because the signal is per trajectory in GRPO and per token in OPD: each trajectory carries one reward, each token carries one teacher score~\cite{agarwal2024onpolicydistillation}. Let
\begin{equation}
    \mu_{\mathcal{G}}
    =
    \frac{1}{m}\sum_{i\in\mathcal{G}}r_i,
    \qquad
    \sigma_{\mathcal{G}}
    =
    \sqrt{
        \frac{1}{m}
        \sum_{i\in\mathcal{G}}
        \left(r_i-\mu_{\mathcal{G}}\right)^2
    }.
\end{equation}
With the group-normalized advantages held constant during differentiation, the group update direction is
\begin{equation}
    \vg_{\mathcal{G}}
    =
    \frac{1}{m}
    \sum_{i\in\mathcal{G}}
    \frac{r_i-\mu_{\mathcal{G}}}
         {\sigma_{\mathcal{G}}+\epsilon}
    \mH_i.
\end{equation}
Although $\mu_{\mathcal{G}}$ and $\sigma_{\mathcal{G}}$ require group completion, the gradient decomposes exactly as
\begin{equation}
    \vg_{\mathcal{G}}
    =
    \frac{1}{m}
    \frac{
        \mG^{(1)}_{\mathcal{G}}
        -
        \mu_{\mathcal{G}}\mG^{(2)}_{\mathcal{G}}
    }{
        \sigma_{\mathcal{G}}+\epsilon
    },
    \qquad
    \mG^{(1)}_{\mathcal{G}}
    =
    \sum_{i\in\mathcal{G}}r_i\mH_i,
    \qquad
    \mG^{(2)}_{\mathcal{G}}
    =
    \sum_{i\in\mathcal{G}}\mH_i.
    \label{eq:thundersync-decomposition}
\end{equation}
\sysname computes $\mH_i$ as soon as trajectory $i$ becomes ready and incrementally accumulates $\mG^{(1)}_{\mathcal{G}}$ and $\mG^{(2)}_{\mathcal{G}}$. When all rewards and source gradients in the group are available, the scalar-weighted combination in Equation~\ref{eq:thundersync-decomposition} gives $\vg_{\mathcal{G}}$. Summing the group directions gives $\vg_k=\sum_{\mathcal{G}\subset\mathcal{B}_k}\vg_{\mathcal{G}}$ for the optimizer barrier. This preserves the batch-barrier update while allowing trajectory backward passes to begin before group completion.

\paragraph{Streaming the OPD gradient.}
In OPD, each turn provides a gradient source once the frozen teacher has scored its action tokens. Let $\mathcal{U}_k$ be the set of turns in $\mathcal{B}_k$, and let $\mathcal{M}_j$ contain the scored action tokens of turn $j$. We normalize the OPD update by the total number of scored tokens in the batch. Define the detached token advantage and the turn's unnormalized update direction as
\begin{equation}
\begin{aligned}
    A_{j,t}
    &= \operatorname{sg}\!\left[
        \log\pi_{\mathrm{T}}(a_{j,t}\mid\vh_{j,t})
        - \log\pi_{\vtheta_k}(a_{j,t}\mid\vh_{j,t})
    \right], \\
    \vg_j
    &= \sum_{t\in\mathcal{M}_j} A_{j,t}
       \left.\nabla_{\vtheta}
       \log\pi_{\vtheta}(a_{j,t}\mid\vh_{j,t})
       \right|_{\vtheta=\vtheta_k},
\end{aligned}
\label{eq:opd-source-gradient}
\end{equation}
where $\pi_{\mathrm{T}}$ is the frozen teacher and $\operatorname{sg}$ denotes stop-gradient. Each advantage depends only on the sampled token, its realized causal context, and the fixed student and teacher policies. \sysname releases the turn-local backward pass as soon as the teacher scores are available, allowing learner computation to overlap with tool execution and subsequent turns. It accumulates unnormalized gradient contributions and completes any deferred gradient propagation through earlier context before final reduction. The batch update direction is
\begin{equation}
    \vg_k
    = \frac{1}{N_k}\sum_{j\in\mathcal{U}_k}\vg_j,
    \qquad
    N_k = \sum_{j\in\mathcal{U}_k}|\mathcal{M}_j|.
    \label{eq:opd-batch-gradient}
\end{equation}
The total scored-token count $N_k$ is fixed only when the logical batch closes, so \sysname defers this scalar normalization while turn-level backward passes proceed. Dividing the accumulated gradient by $N_k$ before clipping and the single optimizer step gives the same mathematical update as batch-synchronous OPD, with every rollout and learner computation bound to $\vtheta_k$. Appendix~\ref{app:update-equivalence} confirms both equivalences numerically on B200 and H200 GPUs.

\section{Experiments}
\subsection{Setup}

\paragraph{Models, datasets and benchmarks.}
\begin{itemize}
    \item \textbf{Models:} We use Qwen-3.8-27B~\citep{qwenteam2026qwen3827b,qwenteam2026qwen38} for full fine-tuning and the separate LoRA configuration (Appendix~\ref{app:lora-efficiency}). The model comparison also includes GLM-4.7-Flash-31B~\citep{glmteam2026glm47flash,glmteam2025glm45}. Appendix~\ref{app:configurations} specifies the Qwen OPD teacher, which remains fixed across execution schedules.
    \item \textbf{Datasets:} We use SWE-smith~\citep{yang2025swesmith} and SETA-Env~\citep{shen2026seta} for post-training, covering repository-level software engineering and broader terminal tasks, respectively. Appendix~\ref{app:datasets} details data revisions, filtering, splits, and benchmark separation.
    \item \textbf{Benchmarks:} We evaluate on SWE-bench Verified~\citep{jimenez2024swebench} and Terminal Bench 4.0~\citep{merrill2026terminalbench}, using the same tasks, agent interface, and inference budget across execution schedules. Terminal Bench 4.0 tasks are excluded from post-training. Appendix~\ref{app:evaluation-setup} specifies success criteria, pass@3 aggregation, and checkpoint comparisons. We also report pass@1 and pass@5 on both benchmarks in Appendix~\ref{app:pass-at-1-and-5}.
\end{itemize}

\paragraph{Workloads and configurations.}
Each GRPO update uses 32 prompts with eight trajectories sampled per prompt, totaling 256 trajectories. Each OPD update uses 32 prompts with four independent trajectories sampled per prompt, totaling 128 trajectories. Both objectives allow up to 300 agentic turns and 1,024 generated tokens per turn. The reference configuration uses a single node with eight NVIDIA B200 GPUs; the hardware comparison also includes a node with eight NVIDIA H200 GPUs. Appendix~\ref{app:configurations} provides the full optimization and generation settings.

\paragraph{Baselines.}
Within each objective, we hold the model, workload, optimizer, precision, and hardware allocation fixed and compare \sysname against three execution schedules:
\begin{itemize}
    \item \textbf{Sync:} A zero-stale baseline that starts learning after the complete rollout batch is available.
    \item \textbf{Async:} A one-update-stale baseline that overlaps learning with the next rollout batch.
    \item \textbf{Fully Async:} Rollout and learning proceed without a batch-level barrier between them. Policy updates can occur during in-flight trajectories, yielding variable policy lag.
\end{itemize}
Sync and Async use the official verl~\citep{sheng2025hybridflow} pipeline for GRPO and the SkyRL~\citep{cao2025skyrl} pipeline for OPD. Sync, Async, and \sysname use vLLM for rollout, with the same chunked prefill and prefix caching settings. We evaluate with the benchmarks' official harnesses; Appendices~\ref{app:generation-settings} and~\ref{app:evaluation-setup} describe the rollout pipelines and harnesses.

\begin{figure}[t!]
  \centering
  \includegraphics[width=\linewidth]{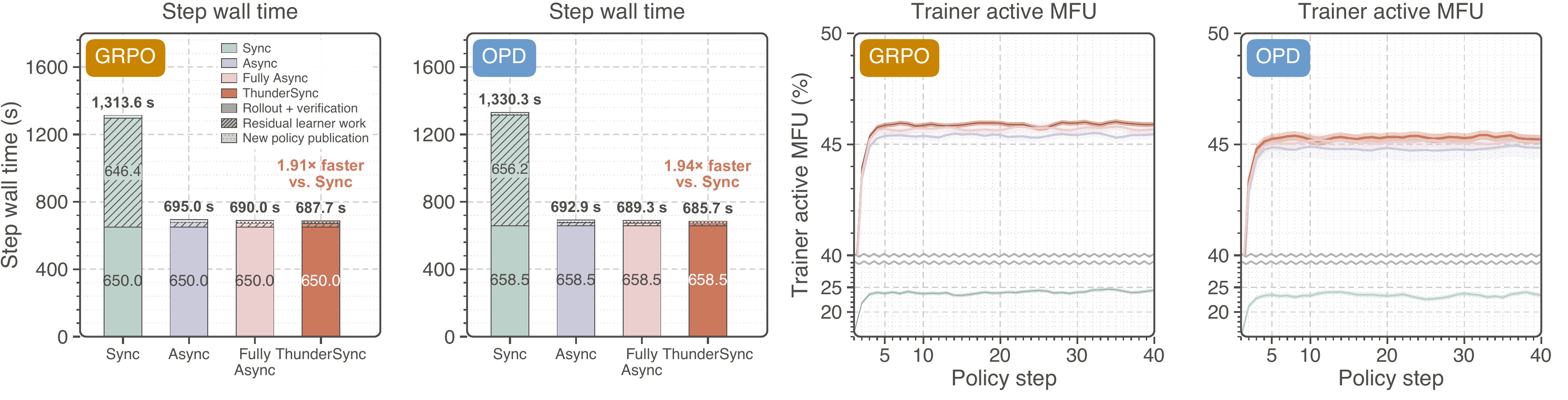}
  \caption{Step wall time (left) and trainer active MFU (right) for GRPO and OPD.}
  \label{fig:trainer-active-mfu}
\end{figure}

\begin{figure}[t!]
  \centering
  \includegraphics[width=\linewidth]{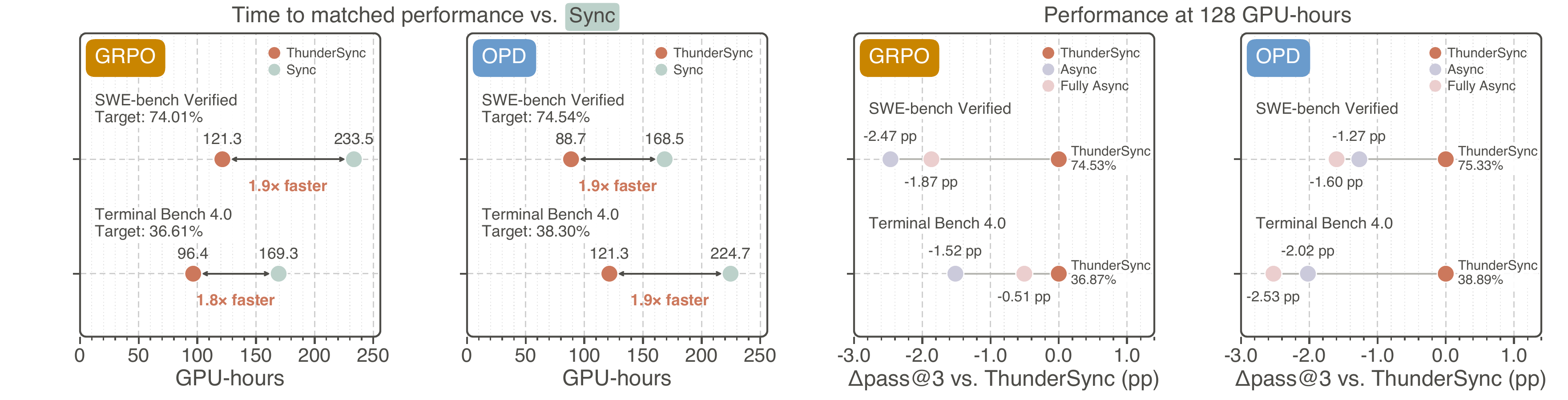}
  \caption{Left: time to matched pass@3. Right: Async and Fully Async relative to \sysname at 128 GPU-hours.}
  \label{fig:time-performance-comparison}
\end{figure}

\begin{figure}[t!]
  \centering
  \includegraphics[width=\linewidth]{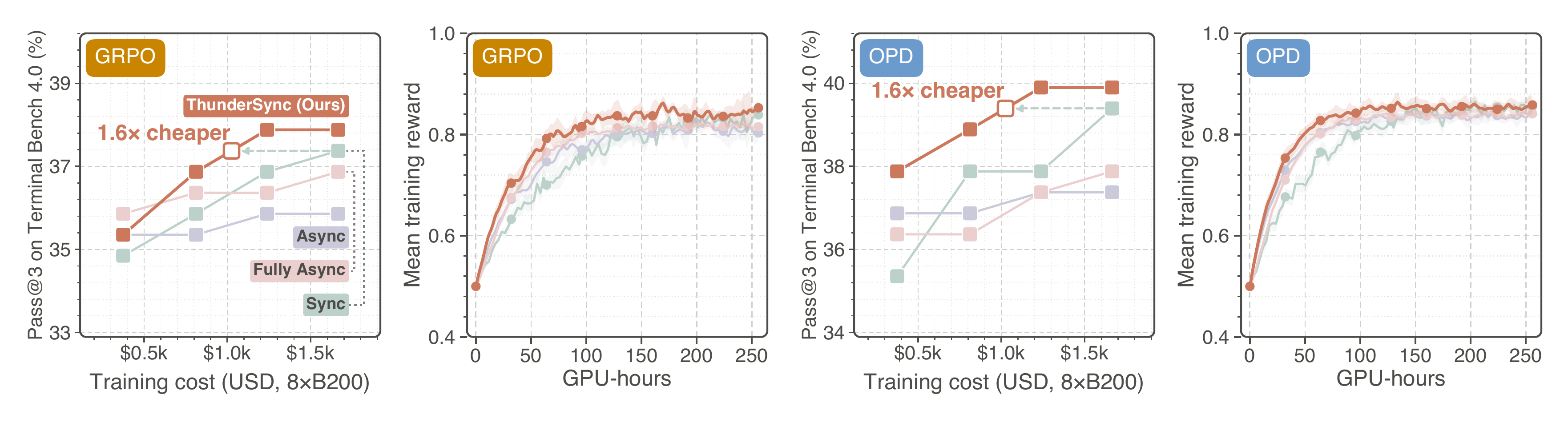}
  \caption{Terminal Bench 4.0 pass@3 vs.\ training cost and mean training reward vs.\ GPU-hours. Arrows mark where \sysname reaches the score of a Sync checkpoint; labels give the cost ratio.}
  \label{fig:terminal-bench-transfer}
\end{figure}

\subsection{Learning efficiency and agent performance}
\paragraph{Time to matched performance.}
\sysname reaches the target pass@3 on SWE-bench Verified $1.9\times$ faster
than Sync under both GRPO and OPD (Figure~\ref{fig:time-performance-comparison}, left).
For example, the GRPO target of 74.01\% is first reached at 121.3 GPU-hours
with \sysname and 233.5 GPU-hours with Sync.
The advantage also holds when training on SETA-Env and evaluating on
Terminal Bench 4.0, whose tasks are excluded from post-training:
\sysname reaches the targets $1.8\times$ faster under GRPO and $1.9\times$
faster under OPD (Figure~\ref{fig:time-performance-comparison}, left;
curves in Figure~\ref{fig:terminal-bench-transfer}).

\paragraph{Performance at a fixed training budget.}
At the same 128-GPU-hour training budget (16 hours on the eight-GPU node),
\sysname achieves higher pass@3 than
Async on both benchmarks (Figure~\ref{fig:time-performance-comparison}, right).
On SWE-bench Verified, \sysname leads by 2.47 percentage points under GRPO
and 1.27 under OPD. On Terminal Bench 4.0, it leads by 1.52 percentage
points under GRPO and 2.02 under OPD.
Each score comes from the latest completed checkpoint available within
the budget; Appendix~\ref{app:evaluation-setup} details the evaluation protocol.

\paragraph{Step wall time.}
\sysname overlaps learner work with rollout (Figure~\ref{fig:trainer-active-mfu}, left).
After 650.0\,s (GRPO) and 658.5\,s (OPD) of rollout and verification, Sync
adds 646.4 and 656.2\,s of learner work, \sysname only 20.5 and 11.6\,s.
With publication, its steps take 687.7 and 685.7\,s, $1.91\times$ and
$1.94\times$ faster than Sync and within Async's seed spread, at zero policy lag.
Over a run, shorter steps fit more trajectory batches into the same GPU-hours,
which is why \sysname reaches each pass@3 level at lower training cost
(Figures~\ref{fig:headline} and~\ref{fig:terminal-bench-transfer}).

\subsection{Efficiency across models and hardware}
\label{sec:efficiency}

The speedup follows how the stages balance on each model and GPU
(Section~\ref{sec:scalability}). On B200, Qwen-3.8-27B is near balance and
reaches $1.91\times$ and $1.94\times$ over Sync under GRPO and OPD.
GLM-4.7-Flash-31B activates about 3B parameters, so its learner work is short,
rollout bounds the step, and the speedup is $1.64\times$ and $1.63\times$
(Table~\ref{tab:model-efficiency}). On H200, the same Qwen workload becomes
learner-bound and reaches $1.63\times$ and $1.67\times$
(Table~\ref{tab:hardware-efficiency}). Each comparison fixes the workload and
GPU allocation across schedules, using Sync on the same hardware as the baseline.

Trainer active MFU divides learner FLOPs by the complete update wall time
and by trainer GPU capacity. The window covers rollout waiting, environment
execution, teacher scoring, learner work, and policy publication, counting
overlap once, so overlapping learner work with rollout raises the metric by
reducing trainer idle time.
\sysname has the highest average active MFU in every configuration, from
18.7\% for GLM to 51.0\% for Qwen on H200. H200's Sync MFU exceeds B200's
because the learner, not rollout, bounds its step, so Sync's learner idles less.
Figure~\ref{fig:trainer-active-mfu} (right) shows the Qwen/B200 curves
stabilizing after startup; Appendix~\ref{app:efficiency-metrics} defines
the FLOP accounting.

\sysname's peak GPU memory exceeds Sync's in every configuration, at most
from 117.0 to 122.8 GB per learner GPU (Qwen, GRPO). These peaks exclude the
pinned host RAM holding the FP32 accumulator and GRPO group statistics
(Appendix~\ref{app:configurations}); rollout and teacher inference use separate GPUs.
Table~\ref{tab:lora-efficiency} reports LoRA (Appendix~\ref{app:lora-efficiency}).

\begin{table}[t!]
  \centering
  \caption{Full fine-tuning efficiency on SWE-smith across models using NVIDIA B200 GPUs.
  GRPO and OPD are compared separately per configuration; for OPD, the model
  is the student.}
  \label{tab:model-efficiency}
  \begingroup
  \normalsize
  \setlength{\tabcolsep}{2pt}
  \renewcommand{\arraystretch}{1.1}
  \resizebox{\linewidth}{!}{%
  \begin{tabular}{@{}lcccc@{\hspace{16pt}}ccc@{}}
    \toprule
    \multirow{2}{*}[-\dimexpr(\aboverulesep+\cmidrulewidth+\belowrulesep)/2\relax]{Method} &
    \multirow{2}{*}[-\dimexpr(\aboverulesep+\cmidrulewidth+\belowrulesep)/2\relax]{\shortstack{Policy lag\\(staleness) $\downarrow$}} &
    \multicolumn{3}{c}{\textbf{GRPO}} &
    \multicolumn{3}{c}{\textbf{OPD}} \\
    \cmidrule(lr){3-5}\cmidrule(l){6-8}
    & & \shortstack{Avg. trainer\\active MFU\\(\%) $\uparrow$} &
    \shortstack{Speedup\\($\times$) $\uparrow$} &
    \shortstack{Trainer peak\\GPU memory\\(GB) $\downarrow$} &
    \shortstack{Avg. trainer\\active MFU\\(\%) $\uparrow$} &
    \shortstack{Speedup\\($\times$) $\uparrow$} &
    \shortstack{Trainer peak\\GPU memory\\(GB) $\downarrow$} \\
    \midrule
    \rowcolor{black!5}[0pt][0pt]
    \multicolumn{8}{@{}l@{}}{\textit{Qwen-3.8-27B}} \\
    \addlinespace[2pt]
    \methoddot{methodSync}Sync &
    \textbf{0} &
    $24.0 \pm 0.3$ &
    $1.00 \pm 0.00$ &
    $\mathbf{117.0 \pm 0.8}$ &
    $23.4 \pm 0.4$ &
    $1.00 \pm 0.00$ &
    $\mathbf{118.0 \pm 0.9}$ \\
    \methoddot{methodAsync}Async &
    1 &
    $45.4 \pm 0.2$ &
    $1.89 \pm 0.03$ &
    $117.2 \pm 1.1$ &
    $44.8 \pm 0.5$ &
    $1.92 \pm 0.04$ &
    $119.3 \pm 1.0$ \\
    \methoddot{methodFullyAsync}Fully Async &
    Variable &
    $45.7 \pm 0.2$ &
    $1.90 \pm 0.03$ &
    $118.6 \pm 1.2$ &
    $45.1 \pm 0.3$ &
    $1.93 \pm 0.04$ &
    $120.1 \pm 1.1$ \\
    \methoddot{methodThunderSync}\textbf{\sysname} &
    \textbf{0} &
    $\mathbf{45.9 \pm 0.1}$ &
    $\mathbf{1.91 \pm 0.04}$ &
    $122.8 \pm 1.3$ &
    $\mathbf{45.3 \pm 0.2}$ &
    $\mathbf{1.94 \pm 0.03}$ &
    $121.3 \pm 1.2$ \\
    \cmidrule(lr){1-8}

    \rowcolor{black!5}[0pt][0pt]
    \multicolumn{8}{@{}l@{}}{\textit{GLM-4.7-Flash-31B}} \\
    \addlinespace[2pt]
    \methoddot{methodSync}Sync &
    \textbf{0} &
    $12.2 \pm 0.1$ &
    $1.00 \pm 0.00$ &
    $\mathbf{124.0 \pm 1.0}$ &
    $11.5 \pm 0.1$ &
    $1.00 \pm 0.00$ &
    $\mathbf{122.1 \pm 0.8}$ \\
    \methoddot{methodAsync}Async &
    1 &
    $19.8 \pm 0.2$ &
    $1.62 \pm 0.03$ &
    $\mathbf{124.0 \pm 1.4}$ &
    $18.4 \pm 0.2$ &
    $1.60 \pm 0.03$ &
    $122.5 \pm 1.3$ \\
    \methoddot{methodFullyAsync}Fully Async &
    Variable &
    $19.9 \pm 0.3$ &
    $1.63 \pm 0.04$ &
    $125.2 \pm 1.3$ &
    $18.5 \pm 0.2$ &
    $1.61 \pm 0.03$ &
    $123.4 \pm 1.2$ \\
    \methoddot{methodThunderSync}\textbf{\sysname} &
    \textbf{0} &
    $\mathbf{20.0 \pm 0.2}$ &
    $\mathbf{1.64 \pm 0.03}$ &
    $129.4 \pm 1.9$ &
    $\mathbf{18.7 \pm 0.1}$ &
    $\mathbf{1.63 \pm 0.02}$ &
    $124.1 \pm 1.5$ \\
    \bottomrule
  \end{tabular}%
  }
  \endgroup
\end{table}

\begin{table}[t!]
  \centering
  \caption{Full fine-tuning efficiency on SWE-smith across GPU platforms with
  Qwen-3.8-27B.}
  \label{tab:hardware-efficiency}
  \begingroup
  \normalsize
  \setlength{\tabcolsep}{2pt}
  \renewcommand{\arraystretch}{1.1}
  \resizebox{\linewidth}{!}{%
  \begin{tabular}{@{}lcccc@{\hspace{16pt}}ccc@{}}
    \toprule
    \multirow{2}{*}[-\dimexpr(\aboverulesep+\cmidrulewidth+\belowrulesep)/2\relax]{Method} &
    \multirow{2}{*}[-\dimexpr(\aboverulesep+\cmidrulewidth+\belowrulesep)/2\relax]{\shortstack{Policy lag\\(staleness) $\downarrow$}} &
    \multicolumn{3}{c}{\textbf{GRPO}} &
    \multicolumn{3}{c}{\textbf{OPD}} \\
    \cmidrule(lr){3-5}\cmidrule(l){6-8}
    & & \shortstack{Avg. trainer\\active MFU\\(\%) $\uparrow$} &
    \shortstack{Speedup\\($\times$) $\uparrow$} &
    \shortstack{Trainer peak\\GPU memory\\(GB) $\downarrow$} &
    \shortstack{Avg. trainer\\active MFU\\(\%) $\uparrow$} &
    \shortstack{Speedup\\($\times$) $\uparrow$} &
    \shortstack{Trainer peak\\GPU memory\\(GB) $\downarrow$} \\
    \midrule

    \rowcolor{black!5}[0pt][0pt]
    \multicolumn{8}{@{}l@{}}{\textit{NVIDIA B200 $\times$ 8}} \\
    \addlinespace[2pt]
    \methoddot{methodSync}Sync &
    \textbf{0} &
    $24.0 \pm 0.3$ &
    $1.00 \pm 0.00$ &
    $\mathbf{117.0 \pm 0.8}$ &
    $23.4 \pm 0.4$ &
    $1.00 \pm 0.00$ &
    $\mathbf{118.0 \pm 0.9}$ \\
    \methoddot{methodAsync}Async &
    1 &
    $45.4 \pm 0.2$ &
    $1.89 \pm 0.03$ &
    $117.2 \pm 1.1$ &
    $44.8 \pm 0.5$ &
    $1.92 \pm 0.04$ &
    $119.3 \pm 1.0$ \\
    \methoddot{methodFullyAsync}Fully Async &
    Variable &
    $45.7 \pm 0.2$ &
    $1.90 \pm 0.03$ &
    $118.6 \pm 1.2$ &
    $45.1 \pm 0.3$ &
    $1.93 \pm 0.04$ &
    $120.1 \pm 1.1$ \\
    \methoddot{methodThunderSync}\textbf{\sysname} &
    \textbf{0} &
    $\mathbf{45.9 \pm 0.1}$ &
    $\mathbf{1.91 \pm 0.04}$ &
    $122.8 \pm 1.3$ &
    $\mathbf{45.3 \pm 0.2}$ &
    $\mathbf{1.94 \pm 0.03}$ &
    $121.3 \pm 1.2$ \\
    \cmidrule(lr){1-8}

    \rowcolor{black!5}[0pt][0pt]
    \multicolumn{8}{@{}l@{}}{\textit{NVIDIA H200 $\times$ 8}} \\
    \addlinespace[2pt]
    \methoddot{methodSync}Sync &
    \textbf{0} &
    $31.3 \pm 0.4$ &
    $1.00 \pm 0.00$ &
    $\mathbf{117.1 \pm 0.9}$ &
    $29.8 \pm 0.4$ &
    $1.00 \pm 0.00$ &
    $118.5 \pm 1.0$ \\
    \methoddot{methodAsync}Async &
    1 &
    $50.4 \pm 0.5$ &
    $1.61 \pm 0.04$ &
    $117.5 \pm 1.2$ &
    $49.2 \pm 0.4$ &
    $1.65 \pm 0.03$ &
    $\mathbf{118.2 \pm 1.1}$ \\
    \methoddot{methodFullyAsync}Fully Async &
    Variable &
    $50.7 \pm 0.4$ &
    $1.62 \pm 0.03$ &
    $118.9 \pm 1.3$ &
    $49.5 \pm 0.5$ &
    $1.66 \pm 0.04$ &
    $119.0 \pm 1.1$ \\
    \methoddot{methodThunderSync}\textbf{\sysname} &
    \textbf{0} &
    $\mathbf{51.0 \pm 0.3}$ &
    $\mathbf{1.63 \pm 0.03}$ &
    $121.9 \pm 1.4$ &
    $\mathbf{49.8 \pm 0.4}$ &
    $\mathbf{1.67 \pm 0.03}$ &
    $120.4 \pm 1.3$ \\
    \bottomrule
  \end{tabular}%
  }
  \endgroup
\end{table}

\subsection{Workload scaling and overlap efficiency}
\label{sec:scalability}

We show how \sysname approaches its $2\times$ speedup bound across stage
balance, trajectory heterogeneity, batch size, and horizon;
Appendix~\ref{app:theory} derives this bound and the release laws.
Every comparison in this section holds the workload and GPU allocation fixed
between Sync and \sysname, so no speedup comes from a change in workload or allocation.

\paragraph{Computation balance.}
Equal stage times offer the largest saving: complete overlap halves
their combined duration, giving $2\times$ speedup when final reduction,
optimizer, and publication overhead is negligible. If one stage dominates, it limits
the saving. Appendix~\ref{app:scalability-grids} sweeps the ratio of
rollout time to learner time at full capacity; one denotes equal times.
We hold this balance below.

\begin{figure}[t!]
  \centering
  \includegraphics[width=\linewidth]{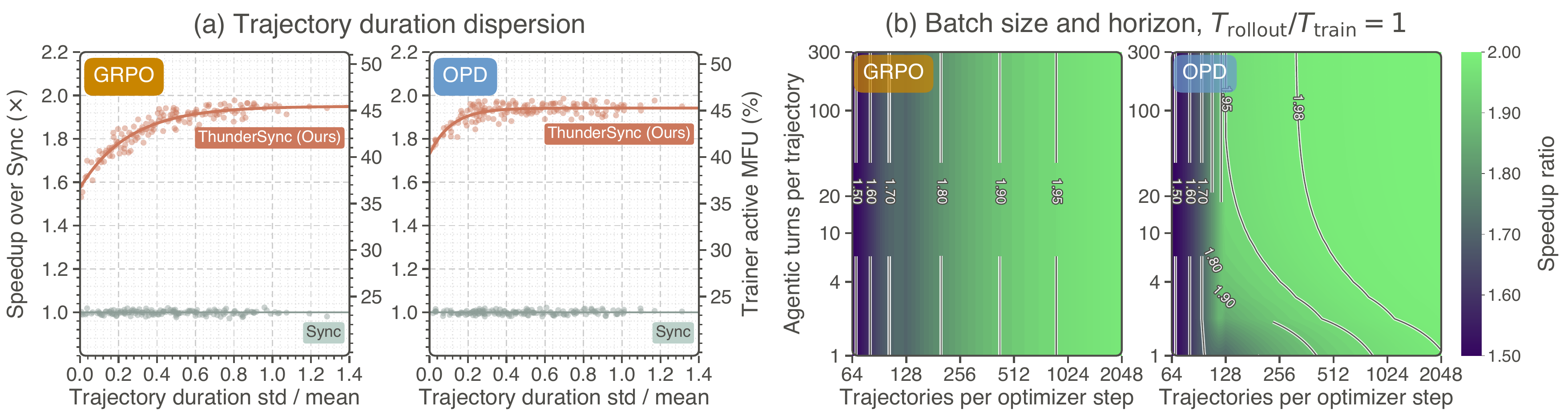}
  \caption{(a) Speedup and trainer active MFU of measured steps versus trajectory duration dispersion, with least-squares fits. (b) Predicted speedup at $T_{\mathrm{rollout}}/T_{\mathrm{train}}=1$.}
  \label{fig:scalability-speedup}
\end{figure}

\paragraph{Trajectory heterogeneity.}
Agentic trajectories vary in length, and Sync's learner waits for the slowest
(Figure~\ref{fig:bubble-motivation}).
\sysname starts a trajectory's backward pass once its reward arrives, so the
learner works while rollout continues. In a separate scaling setup, over 150 measured steps per objective at
balance (Figure~\ref{fig:scalability-speedup}a), GRPO's speedup rises from
$1.58\times$ at near-equal durations, where only rollout waves overlap, to
$1.95\times$ as the spread grows. OPD releases work per turn and stays at
$1.73$ to $1.94\times$.

\paragraph{Batch and horizon.}
Figure~\ref{fig:scalability-speedup}b varies trajectories per optimizer step
and agentic turns per trajectory. Rollout capacity is fixed at 128
trajectories, so larger batches run in successive waves. \sysname processes ready
work from earlier waves while later waves run. At balance, only the final
wave's learner work remains after rollout ends, forming the
\emph{learner tail}. As the batch grows, the tail and policy
publication take a smaller share of the update, so the predicted
speedup approaches $2\times$.

With the completion-time distribution held fixed, GRPO's predicted speedup
does not change with turn count, because a trajectory's work is released only
when its reward arrives. For OPD, more
turns spread releases and shorten the tail, with diminishing gains limited by
deferred work (Appendix~\ref{app:scalability-release-laws}).

\begin{figure}[t!]
  \centering
  \includegraphics[width=\linewidth]{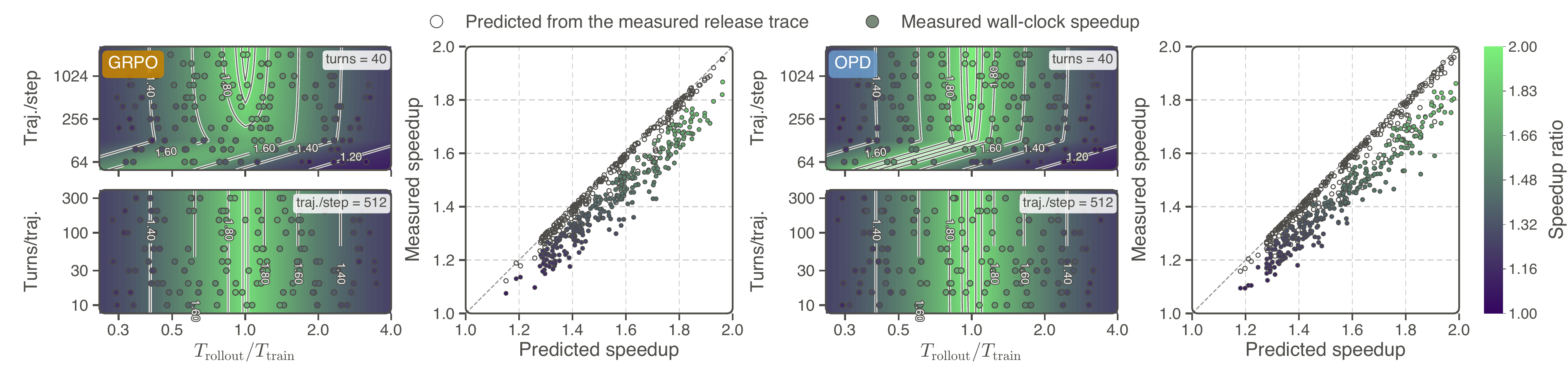}
  \caption{Speedup over Sync. Contours show the analytical prediction across batch size (upper) and horizon (lower). Scatter panels pool both sweeps and plot this prediction against predictions from measured release traces (open circles) and measured wall-clock speedups (filled circles).}
  \label{fig:speedup-prediction}
\end{figure}

\paragraph{Prediction and overhead.}
For Figure~\ref{fig:speedup-prediction}, we measured 588 paired configurations
across both objectives, varying batch size at 40 turns and horizon at
512 trajectories while sweeping the rollout-to-learner time ratio.
These scaling experiments use a separate setup from the learning-curve
runs; Appendix~\ref{app:scalability-figure-inputs} specifies allocation and concurrency.
Predictions from measured release traces differ from
analytical predictions by 1.6\% on average and at most 10.6\%; measured
wall-clock speedups average 6.5\% below the trace predictions. These gaps
describe release-model error and the remaining runtime overhead, respectively
(Appendix~\ref{app:scalability-figure-inputs}).

\section{Conclusion}
We presented \sysname, which overlaps rollout and learning with zero off-policy staleness. It streams ready gradients at a fixed policy snapshot and takes one optimizer step after the batch closes, preserving the synchronous GRPO and OPD updates. Gradient sufficient statistics let GRPO backward passes start before all group rewards arrive, and teacher-scored turns let OPD learning proceed during tool execution. Step wall time speedup can approach $2\times$ when rollout and learner work are balanced and the learner tail and update overhead are negligible. Agent training at frontier scale moves only as fast as experiments turn into evidence. By recovering idle learner time without changing the objective, \sysname shortens that loop, so the same compute answers more research questions. Further discussion, including limitations, appears in Appendix~\ref{app:discussion}.

\clearpage
\printbibliography

\clearpage
\appendix
\appendixpage
\startcontents[sections]
\printcontents[sections]{l}{1}{\setcounter{tocdepth}{2}}
\pagebreak
\section{Limitation and further discussion}
\label{app:discussion}

\sysname rests on one premise: learner work should begin as soon as the
objective's dependencies allow. How much this saves is therefore set as much
by the workload as by the scheduler. Objectives whose gradient inputs resolve
early, such as turn-level OPD, leave more learner work to overlap with rollout
than objectives that wait on group-level signals, and the benefit is largest
when rollout and learner work take comparable time. Starting work early also
means keeping partial gradient state until each group or batch closes, which
exchanges trainer memory for wall time. We regard these dependencies as a
design space. Finer readiness units and admission that weighs memory against
remaining service would move more learner work into the rollout window, and
identifying readiness units for other objectives would extend exact gradient
streaming beyond GRPO and OPD. Scaling the same principle to multi-node
training and a broader set of agentic workloads is a natural next step.
\section{Efficiency measurement}
\label{app:efficiency-metrics}

\subsection{Trainer active MFU}

Throughout this paper, trainer active MFU denotes the learner's model
FLOPs divided by the complete update wall time and the peak capacity of
the trainer GPUs. We use this label for the step-averaged trainer
metric, whose time window includes rollout waiting and trainer idle
intervals.

For update $k$, let $F_k^{\mathrm{train}}$ be the model FLOPs required by
the learner, $T_k^{\mathrm{step}}$ its complete update wall time, and $P_d$
the dense-BF16 peak rate used for trainer GPU $d$. We define
\begin{equation}
    \operatorname{MFU}^{\mathrm{active}}_k
    = \frac{F_k^{\mathrm{train}}}
    {T_k^{\mathrm{step}}\sum_{d\in\mathcal{D}_{\mathrm{train}}}P_d}.
    \label{eq:trainer-active-mfu}
\end{equation}
The capacity denominator includes the trainer GPUs, with the same peak-rate convention for all methods within a hardware configuration. Figure~\ref{fig:trainer-active-mfu} and Tables~\ref{tab:model-efficiency}, \ref{tab:hardware-efficiency}, and~\ref{tab:lora-efficiency} use this definition.

The denominator uses the same wall-clock intervals as step wall time
below. For Sync and \sysname, an interval begins with the update's rollout
and ends when the rollout replicas acknowledge the published policy.
It includes environment and reward processing, teacher scoring, all
learner work, final reduction, optimizer execution, and publication.
Waiting for a source to become ready remains included. For
Async, we use complete steady-state update cycles between successive
policy-publication acknowledgements, counting concurrent stages once.
For \sysname, deferred context gradients and shared-prefix backward must
finish within the measured update.

The FLOPs numerator follows the model architecture and the contexts processed by the learner. Context tokens contribute computation even when their loss mask is zero. For LoRA, the count includes forward computation and backward propagation through frozen weights, as well as adapter gradients. Activation-checkpoint recomputation contributes to elapsed time without being credited as additional model FLOPs. Rollout generation and teacher scoring contribute to elapsed time, but their FLOPs and GPU capacity are excluded from this trainer-side metric.

\paragraph{Operator accounting.}
For a matrix product with dimensions $(m\times k)(k\times n)$, the
counter credits $2mkn$ FLOPs. A linear layer processing $T$ tokens with
input/output widths $d_{\mathrm{in}},d_{\mathrm{out}}$ therefore costs
$2T d_{\mathrm{in}}d_{\mathrm{out}}$ per forward pass. When both input and
weight gradients are required, backward adds two such products. A frozen
LoRA base omits the weight-gradient product but retains the input-gradient
product wherever gradients propagate through it. The rank-$r$ adapter's
forward and backward products are counted separately.

The ledger sums executed attention projections, query-key and attention-value
products, dense and recurrent blocks, and vocabulary projections using
their actual tensor shapes. For GLM, routed-expert products use the
number of tokens dispatched to each expert; shared experts and routing
computation are included separately. Neither total stored parameters nor
the nominal active-parameter count alone is an adequate FLOP counter.
Padding that reaches a kernel counts as executed work. Each record keeps
useful model work separate from checkpoint recomputation, so the latter
increases elapsed time without inflating the credited numerator.

\subsection{Average trainer active MFU}

For a reporting window of $N$ completed updates after warmup, average trainer active MFU is
\begin{equation}
    \overline{\operatorname{MFU}}^{\mathrm{active}}
    = \frac{1}{N}\sum_{k=1}^{N}
    \operatorname{MFU}^{\mathrm{active}}_k.
    \label{eq:average-trainer-active-mfu}
\end{equation}
Each term uses the learner FLOPs and complete wall time of the same
update cycle. With fixed learner work and trainer capacity, shorter
updates increase trainer active MFU. Overlapping learner computation
with rollout can therefore raise this metric without changing kernel
throughput. Within one configuration, all schedules process the same
learner FLOPs, so the MFU ratio to Sync tracks the step wall time
speedup; across models and hardware, learner FLOPs differ.

\paragraph{Profiling window and aggregation.}
We completed 40 updates for each of seeds 42, 43, and 44, discarded
updates 1--4 as startup, and profiled updates
5--40 ($N=36$). Startup includes initial compilation and allocator/cache
growth; it is distinct from learning-rate warmup. All schedules retain
the same number of complete steady-state intervals, with extra pipeline
startup/drain work recorded outside the Async measurement window.
The wall-time denominator uses a common host timebase for complete
update cycles. Gaps with no admitted learner work remain included;
we do not replace this window with a union of active kernel intervals.

CUDA events recorded device service without synchronizing after every
turn. We collected detailed operator counts in a separate untimed trace
pass with the same workload and tensor shapes. Within each
seed, MFU is averaged over the 36 updates and speedup is the ratio of mean
step times. We report the mean and, where shown, sample standard deviation across the
three seed-level summaries, rather than treating 108 correlated updates
as independent runs.

\subsection{Step wall time speedup}

Step wall time is the elapsed wall-clock time per completed training update. The interval includes rollout, environment and reward processing, learner work, final reduction, and policy publication. It ends when the rollout replicas acknowledge the updated policy. For $N$ completed updates, let $\bar{T}_{\mathrm{method}}^{\mathrm{step}}$ be elapsed time divided by $N$. Speedup for each objective and configuration is
\begin{equation}
    \operatorname{Speedup}_{\mathrm{method}}
    = \frac{\bar{T}_{\mathrm{Sync}}^{\mathrm{step}}}
    {\bar{T}_{\mathrm{method}}^{\mathrm{step}}},
\end{equation}
where both times use the same workload, hardware allocation, and number of completed updates after warmup. This is a ratio of mean step wall times, not a mean of per-step ratios or a ratio of kernel execution times. For Async, the window covers completed steady-state update cycles with the declared one-update policy lag; overlapping stages contribute once to elapsed time.

\subsection{Hardware allocation and memory}

Each eight-GPU configuration assigns four GPUs to the learner. GRPO uses
the remaining four for rollout; OPD uses two for rollout and two for the
teacher. \sysname replicates parameters, shards optimizer state, and
accumulates gradients on the host. Appendix~\ref{app:configurations} gives
the precision and execution settings. The dense-BF16 peak rates used in
the MFU denominator are 2,250 TFLOP/s per B200 GPU and 989.5 TFLOP/s per
H200 GPU.

We report two kinds of memory. GPU memory is the HBM on each learner GPU;
host memory is the CPU RAM of the node.
Tables~\ref{tab:model-efficiency} and~\ref{tab:hardware-efficiency}
report trainer peak GPU memory per GPU. The reported peak is the largest
among the learner GPUs during an update, including gradient accumulation,
distributed reduction, and optimizer execution. It covers resident model
and optimizer state, activations, gradient buffers, and temporary storage
on those GPUs. Rollout KV caches and teacher memory belong to their
separate GPU pools. The serving memory fraction of 0.65 controls
the rollout executors' memory budget.

We reset each learner process's CUDA peak-memory counters at the
start of the profiled update, after the persistent model and optimizer
state is resident, and read them after all backward, reduction, and
optimizer work has completed. We report
\ensuremath{\mathtt{torch.cuda.max\_memory\_allocated()}} as the tensor
allocation peak, taking the maximum across learner ranks and profiled
updates within each seed, and convert bytes to decimal GB ($10^9$ bytes).
We record reserved allocator memory and device-wide memory from NVML
separately, since they include different allocations.\footnote{The CUDA
memory APIs and the distinction between allocated and reserved memory are
documented in the \href{https://pytorch.org/docs/stable/notes/cuda.html\#cuda-memory-management}{PyTorch memory-management notes}.}
Deferred-state buffers that stay on the learner GPUs count toward
\sysname's GPU peak. The FP32 batch accumulator and the GRPO group
statistics sit in pinned host memory, so this counter does not include
them; Appendix~\ref{app:configurations} bounds their host-memory size.
A model-state size or a serving-memory limit is not a measured
peak. We use the same collection procedure for the full fine-tuning and
LoRA comparisons.

\section{Training data and benchmark separation}
\label{app:datasets}

\subsection{Data sources and versions}
The software-engineering workload uses SWE-smith, and the terminal workload
uses SETA-Env. These are separate post-training workloads, evaluated on
SWE-bench Verified and Terminal Bench 4.0, respectively. Table~\ref{tab:data-revisions}
fixes the latest public revisions inspected on September 6, 2026.
Revisions stay fixed throughout a comparison, including benchmark task
definitions and evaluation resources.

\begin{table}[ht]
  \centering
  \caption{Pinned data sources. Sizes describe the inspected source pools
  before the filtering and development split in Table~\ref{tab:data-splits}.}
  \label{tab:data-revisions}
  \small
  \setlength{\tabcolsep}{5pt}
  \begin{tabular}{@{}llrl@{}}
    \toprule
    Collection & Role & Tasks & Revision or release \\
    \midrule
    SWE-smith Python & Post-training & 50,908 & \ensuremath{\mathtt{77cab9055d42}} \\
    SETA-Env & Post-training & 4,567 & \ensuremath{\mathtt{3c3bc8975b05}} \\
    SWE-bench Verified & Evaluation & 500 & \ensuremath{\mathtt{78f471bf655a}} \\
    Terminal Bench 4.0 & Evaluation & 66 & \ensuremath{\mathtt{v4.0.0\ /\ 452bf305c6da}} \\
    \bottomrule
  \end{tabular}
\end{table}

\paragraph{SWE-smith.}
SWE-smith constructs executable repository-repair tasks by introducing
bugs into working codebases and identifying tests that fail after the
change~\citep{yang2025swesmith}. Its construction methods include language-model
rewrites, procedural edits, combined bugs, and pull-request mirroring.
The original paper reports 50,137 instances from 128 repositories. The
selected, maintained Python release contains 50,908 instances from 131
repositories.\footnote{The selected source is
\href{https://huggingface.co/datasets/SWE-bench/SWE-smith-py/tree/77cab9055d42ab4a5c25c89a8f937096db13558e}{the maintained SWE-smith Python release}.
The older monolithic \ensuremath{\mathtt{SWE\text{-}bench/SWE\text{-}smith}} release has different
contents and directs users to the language-specific collections.}
Each record supplies a problem statement, a repository environment, a
bug-inducing patch, and test identifiers. The environment setup must apply
this patch before agent interaction; its direction is the opposite of the
solution patch in SWE-bench. Existing repository tests can be visible during
SWE-smith training, so training reward and performance on held-out repair
tasks measure different things.

\paragraph{SETA-Env.}
SETA builds terminal environments from community questions and answers,
Kaggle notebooks, and natural-language shell-command examples.
SETA-Synth creates tasks with executable environments and verifiers;
SETA-Evol changes their difficulty or context~\citep{shen2026seta}.
The inspected release has 3,255 synthesized and 1,312 evolved task
directories, matching the 4,567 tasks reported in its paper and dataset
card.\footnote{The source is
\href{https://huggingface.co/datasets/camel-ai/SETA-Env/tree/3c3bc8975b05826bf41769bd6b2da76ad1b6dd42}{\ensuremath{\mathtt{camel\text{-}ai/SETA\text{-}Env}}}.}
Each task includes instructions, an environment definition, a reference
solution, and executable tests. SETA's construction procedure checks a
no-op solution and an oracle solution, then audits systematic rollout
failures for task-design flaws. These upstream checks describe dataset
construction; they do not replace validation of the environments used in
\sysname runs.

\subsection{Filtering and development splits}
\label{app:data-filtering}
The static SWE-smith filter requires a nonempty instance identifier,
problem statement, repository identifier, and environment image, and
retains one record per instance identifier. This removes 11,437 records,
all with empty problem statements. The remaining 39,471 instances cover
131 repositories. Before comparing repository identities, the filter maps
SWE-smith mirror names back to their original owner/repository names and
normalizes case. None of these repositories matches the 12 repositories
in the pinned SWE-bench Verified set.

For SETA, the static filter requires \ensuremath{\mathtt{task.toml}}, instructions,
an environment Dockerfile, a test entrypoint, and a reference-solution
entrypoint. All 4,567 task directories pass this check.
A source-family identifier removes the evolution suffix from an evolved
task's identifier. Tasks in the same source family stay in one split, including evolved
tasks whose original is absent from the release.

The development split contains 5\% of groups, rounded up: repositories for
SWE-smith and source families for SETA. Groups are ordered by the SHA-256
digest of the normalized group identifier; the first groups form the
development set. The remaining
groups form the training set. This produces Table~\ref{tab:data-splits}.
Task lists and their digests define the split independently of model
performance or trajectory completion time.
This five-percent group split is a \sysname protocol choice, not an official
development split supplied by either dataset.

\begin{table}[ht]
  \centering
  \caption{Statically constructed cohorts for the declared protocol,
  before prompt-length checks, container validation, and semantic-overlap review.}
  \label{tab:data-splits}
  \small
  \begin{tabular}{@{}lrr@{}}
    \toprule
    Stage & SWE-smith Python & SETA-Env \\
    \midrule
    Source tasks & 50,908 & 4,567 \\
    Removed by static field/file checks & 11,437 & 0 \\
    Eligible tasks & 39,471 & 4,567 \\
    Split groups & 131 repositories & 3,474 source families \\
    Training tasks & 37,084 & 4,333 \\
    Development tasks & 2,387 & 234 \\
    Training/development groups & 124 / 7 & 3,300 / 174 \\
    \bottomrule
  \end{tabular}
\end{table}

\paragraph{Executable validation.}
For SWE-smith, we ran the validation harness on the original repository,
then applied the bug-inducing patch and tested it again. Its collection
rule requires at least one repair test and one retained passing test;
evaluation with the gold repair provides a separate harness sanity
check.\footnote{See the \href{https://swesmith.com/guides/harnesses/}{SWE-smith validation and evaluation guide}.}
For SETA, we started a fresh environment for each of the no-op and
reference solutions, recorded the verifier output, and required the
reference solution to pass without the no-op already
solving the task. These checks validate the environment and reward signal,
not the trained policy.

The run manifest records the task identifier, source revision, image
digest, verifier revision, validation outcome, and any exclusion reason.
Prompt-length filtering uses the pinned student tokenizer and the actual
serialized system, tool, and task prompt before rollout. A failing task
is removed from every schedule's cohort before shuffling; its removal
updates the task count and list digest. Table~\ref{tab:data-splits} reports
the static cohort construction; the execution manifest records subsequent
validation exclusions and the final task lists used by each comparison.

A task's difficulty, a zero-variance GRPO group, or a long trajectory does
not trigger exclusion. Although campaign-specific SETA launchers can
select benchmark tasks or discard uniform-reward groups,\footnote{See the
\href{https://github.com/camel-ai/seta/blob/e4715b01174e6c9503fc46120d81dd692ced75e6/scripts/miles/run_glm47_flash_seta_session_server.py}{versioned SETA GLM training launcher}.}
our comparison retains those groups and excludes Terminal Bench tasks
from training. Within a seed, schedules draw prompts from the same shuffled
training list without replacement, reshuffling when it is exhausted.
The two datasets are not mixed, and development tasks contribute no
training gradients.

\subsection{Scope of benchmark separation}
Excluding benchmark task identifiers is only one check on data separation.
The SWE-smith paper describes using a SWE-bench Verified problem statement
as a style demonstration during issue generation (Appendix D of
\citealp{yang2025swesmith}). Repository disjointness therefore does not
establish that its construction was independent of benchmark text.
For SETA, checks must cover instructions, reference solutions, tests, and
source ancestry, including relationships to earlier Terminal Bench releases.
The static checks above do not establish semantic decontamination.
The overlap review first matches task IDs, canonical repository or source
IDs, and hashes of whitespace-normalized instructions, solutions, and
tests. Candidate paraphrases and evolved descendants then require source
inspection. A confirmed benchmark derivative excludes its entire source
family before splitting; ambiguous cases remain recorded rather than
being certified clean. Counts of candidates, confirmed overlaps, and
exclusions accompany the final cohort manifest.

The protocol excludes benchmark tasks, solutions, verifier outputs, and
identified derivatives from training, prompts, and development selection.
Upstream data generation or model pretraining may still expose the student
or teacher to benchmark content. Results therefore compare schedules with
matched data and models, without establishing that every task was unseen
throughout model development.

\section{Optimization and generation settings}
\label{app:configurations}

\subsection{Logical batches and optimization}
Table~\ref{tab:optimization-settings} gives shared optimization settings
for Sync, Async, and \sysname within each objective. The main
configuration updates the student's language-model parameters. The text-only
agent interface does not use Qwen's vision components. The LoRA
configuration is separate from full fine-tuning.

\begin{table}[ht]
  \centering
  \caption{Declared optimization settings. Each column is a separate
  comparison among execution schedules. The learning rate for the
  separate LoRA configuration is $10^{-5}$.}
  \label{tab:optimization-settings}
  \small
  \setlength{\tabcolsep}{5pt}
  \begin{tabular}{@{}lll@{}}
    \toprule
    Setting & GRPO & OPD \\
    \midrule
    Prompts per optimizer update & 32 & 32 \\
    Trajectories per prompt & 8 & 4 independent samples \\
    Trajectories per optimizer update & 256 & 128 \\
    Optimizer & AdamW & AdamW \\
    Full fine-tuning learning rate & $10^{-6}$ & $10^{-6}$ \\
    Learning-rate schedule / warmup & Constant / none & Constant / none \\
    Adam $(\beta_1,\beta_2)$ / $\epsilon$ & $(0.9,0.999)$ / $10^{-8}$ & $(0.9,0.999)$ / $10^{-8}$ \\
    Weight decay & 0 & 0 \\
    Global gradient-norm clipping & 1.0 & 1.0 \\
    Optimizer steps per logical batch & 1 & 1 \\
    Additional PPO epochs & 0 & 0 \\
    Reference-policy KL / entropy bonus & 0 / 0 & 0 / 0 \\
    Loss-bearing tokens & Generated assistant tokens & Generated assistant tokens \\
    Reduction & Mean over trajectories & Global scored-token mean \\
    Advantage stabilization $\epsilon$ & $10^{-6}$ & Not applicable \\
    Student forward precision & BF16 & BF16 \\
    Gradient accumulation / Adam state & FP32 (host) / BF16 & FP32 (host) / BF16 \\
    Training seeds & 42, 43, 44 & 42, 43, 44 \\
    \bottomrule
  \end{tabular}
\end{table}

The full fine-tuning learning rate, Adam moments, clipping threshold, and
temperature-one sampling follow conventional GRPO settings, including
the corresponding TRL defaults. They are fixed across schedules rather
than selected separately for each method.\footnote{See the
\href{https://huggingface.co/docs/trl/en/grpo_trainer}{TRL GRPO configuration}.
The table does not list all TRL, SWE-smith, or SETA defaults.}
The constant learning rate, trajectory-mean GRPO reduction, and single
logical update are explicit \sysname choices; adopting an entire framework's
default configuration would not preserve the objective defined here.
We save the resolved configuration with each run's model and data revisions.
For GRPO, let $L_i$ be trajectory $i$'s scored-token count. Setting
$w_{i, t}=1/(32L_i)$ in Section~\ref{sec:gradient-streaming} makes the sum
of group directions a mean over the 256 trajectory-normalized directions.
Rewards are standardized within each eight-trajectory group using the
population standard deviation. Uniform-reward groups contribute zero
policy gradient and retain their place in the batch and timing record.
The group-normalized advantages are held constant during differentiation,
and each logical batch produces one optimizer step.

OPD applies the sampled-token log-probability difference in
Equation~\ref{eq:opd-source-gradient}, with both distributions evaluated
at temperature one. Four trajectories per prompt provide independent
samples; OPD has no group-relative reward normalization.
Accumulation uses the global scored-token denominator in
Equation~\ref{eq:opd-batch-gradient}. Both objectives normalize
and reduce gradients across ranks before global clipping and the single
optimizer step. Prompt and observation tokens supply causal context without direct loss.

For the Qwen LoRA configuration, the specified rank is 32, scaling parameter
$\alpha=64$, dropout is zero, and the base weights remain frozen. The
targets are \ensuremath{\mathtt{q\_proj}}, \ensuremath{\mathtt{k\_proj}}, \ensuremath{\mathtt{v\_proj}},
\ensuremath{\mathtt{o\_proj}}, \ensuremath{\mathtt{gate\_proj}}, \ensuremath{\mathtt{up\_proj}},
\ensuremath{\mathtt{down\_proj}}, \ensuremath{\mathtt{in\_proj\_qkv}}, \ensuremath{\mathtt{in\_proj\_z}},
\ensuremath{\mathtt{in\_proj\_b}}, \ensuremath{\mathtt{in\_proj\_a}}, and \ensuremath{\mathtt{out\_proj}}.
Batch sizes, generation limits, and evaluation settings remain those of
the corresponding objective.

\subsection{Agent interaction and generation}
\label{app:generation-settings}
\paragraph{Rollout pipelines.}
We use verl's official Agent Loop for GRPO and SkyRL's official
multi-turn generation pipeline, through $\mathtt{SkyRLGymGenerator}$
and SkyRL-Gym, for OPD. Both use vLLM as the rollout engine. The
agent loop alternates model generation with shell-tool execution in
the task environment and retains the resulting history for the next
turn. Independent trajectories run concurrently, so a tool call waiting
for its result does not block generation for other trajectories.\footnote{See
the \href{https://verl.readthedocs.io/en/latest/advance/agent_loop.html}{verl Agent Loop}
and \href{https://docs.skyrl.ai/docs/examples/multi_turn_text2sql}{SkyRL multi-turn rollout documentation}.}
The same agent interface and vLLM serving settings apply to Sync, Async,
and \sysname within each objective.

\begin{table}[ht]
  \centering
  \caption{Generation settings shared across schedules. Training and
  evaluation use the same per-attempt token and turn limits.}
  \label{tab:generation-settings}
  \small
  \begin{tabular}{@{}ll@{}}
    \toprule
    Setting & Value \\
    \midrule
    Maximum agentic turns per trajectory & 300 \\
    Maximum generated tokens per turn & 1,024 \\
    Maximum total generated tokens per trajectory & 16,384 \\
    Maximum initial prompt tokens & 16,384 \\
    Maximum total context tokens & 131,072 \\
    Maximum retained observation per tool call & 4,096 tokens (head and tail equally) \\
    Sampling temperature / top-$p$ / top-$k$ & 1.0 / 1.0 / disabled \\
    Repetition / presence / frequency penalties & 1.0 / 0 / 0 \\
    Thinking mode & Enabled; reasoning tokens count toward the limits \\
    Tool calls per agent turn & At most one \\
    Ordinary tool-command timeout & 120 seconds \\
    Student rollout weights / KV cache & BF16 / BF16 \\
    Student serving memory fraction & 0.65 \\
    Maximum active sequences per rollout replica & 16 \\
    Chunked prefill / prefix caching & Enabled / enabled \\
    Speculative decoding & Disabled \\
    Evaluation attempts per task and checkpoint & 3 (5 in Appendix~\ref{app:pass-at-1-and-5}) \\
    \bottomrule
  \end{tabular}
\end{table}

\paragraph{Shared serving optimizations.}
All schedules enable vLLM's chunked prefill and automatic prefix caching,
and disable speculative decoding. Chunked prefill batches prompt chunks
with decode requests. Prefix caching reuses matching cached prefixes,
including retained conversation history across turns when the cache is
available on the serving replica. Policy publication invalidates
policy-dependent caches before requests use the new weights. These
serving optimizations are common to the comparison; \sysname changes when
ready learner work is admitted.\footnote{See vLLM's
\href{https://docs.vllm.ai/en/stable/configuration/optimization/}{chunked prefill}
and \href{https://docs.vllm.ai/en/latest/features/automatic_prefix_caching/}{automatic prefix caching documentation}.}

The limits in Table~\ref{tab:generation-settings} are independent ceilings.
In particular, 300 turns do not imply an
allowance of $300\times1{,}024$ generated tokens once the trajectory-level
limit is reached. Initial prompts exceeding their limit are recorded
before training and excluded consistently across schedules. During a
trajectory, reaching the generation or context limit ends the attempt;
previous turns are not removed to create additional context capacity.
The agent retains the first and last 2,048 tokens of longer tool outputs,
with the same truncation rule in every arm. An unparsable action consumes
its turn and returns a fixed parser-error observation. Each independent
attempt starts from a fresh task environment.

The agent interface supplies the task instruction and a common shell tool,
with fixed system and tool descriptions within a model configuration.
There is no retrieval from benchmark solutions or evaluator-controlled
files. The teacher scores the student's serialized causal history,
including its retained observations, rather than a separately rewritten
conversation. Qwen reasoning history is preserved in that history.
Sampling seeds are assigned by run seed, optimizer update, task identifier,
trajectory index, and turn index, independently of arrival order.
Matching these seeds does not imply identical trajectories once the
policies differ, particularly for Async.

\subsection{Teacher and hardware allocation}
The teacher for Qwen-3.8-27B is
Qwen/Qwen3.8-Flash-Next-FP8, revision
\ensuremath{\mathtt{236dfdf28582}}; the student is Qwen/Qwen3.8-27B,
revision \ensuremath{\mathtt{1d4bf0f2ff60}}.
The identical tokenizer gives token-level correspondence for OPD. The
teacher remains frozen, uses its released FP8 quantization configuration
on both GPU types, and supplies log probabilities only for the sampled
assistant tokens. Log-probability normalization and score transport use
FP32; this does not undo approximation introduced by FP8 teacher weights.
Teacher scoring uses the same causal prefix and scoring implementation
across schedules, with work becoming eligible after each completed turn.

The hardware configurations are one node with eight B200 GPUs and one
with eight H200 GPUs. The declared allocation reserves four GPUs for the
student learner. GRPO uses the remaining four for rollout replicas; OPD
uses two for rollout replicas and two for the teacher, with teacher tensor
parallelism of two. Student replicas each use one GPU. \sysname
replicates the BF16 student on each of the four learner GPUs, shards the
AdamW state across them, and accumulates gradients in FP32 in host memory,
with activation checkpointing; this lets each rank release sources
independently. Under GRPO, each open group also keeps its two FP32
statistics $\mG^{(1)}_{\mathcal{G}}$ and $\mG^{(2)}_{\mathcal{G}}$ in pinned
host memory; they are allocated with the group's first source and freed
when the group closes. An optional admission cap $k$ defers the backward
of a trajectory from a new group while $k$ groups on its rank hold
statistics, so only its tokens wait. The statistics and the batch
accumulator then occupy at most $4(2k+1)P$ bytes of host memory per rank
for $P$ parameters, regardless of the number of prompts per batch. We set
$k=1$, which bounds the statistics and batch accumulator at 1.30\,TB for
Qwen and 1.49\,TB for GLM across the four learner ranks. The Sync and Async baselines use the same GPU allocation and the optimization
and precision settings of Table~\ref{tab:optimization-settings}.
With 16 active sequences per replica, the reference allocation permits
up to 64 concurrently scheduled generation sequences for GRPO and 32
for OPD. These limits govern vLLM generation, separately from the number
of trajectories in a logical update or waiting on tools. Environment
and verifier resource limits and concurrency settings are fixed across
schedules. The resolved launch configurations will accompany the code
release. The workload-scaling experiments use a separate allocation and
trajectory-capacity setup (Appendix~\ref{app:scalability-figure-inputs}).

The Qwen teacher's released tensors occupy approximately 185.5\,GB before
quantization scales, caches, and runtime buffers. Its two-GPU allocation
and scoring procedure remain fixed across schedules.

\paragraph{GLM configuration.}
\label{app:glm-settings}
The GLM student is \ensuremath{\mathtt{zai\text{-}org/GLM\text{-}4.7\text{-}Flash}},
revision \ensuremath{\mathtt{7dd20894a642}}, with about 31.22 billion stored
parameters and approximately three billion active parameters per token.
Its released configuration has 47 decoder layers, 64 routed experts per
MoE layer, and four selected experts per token.\footnote{See the pinned
\href{https://huggingface.co/zai-org/GLM-4.7-Flash/tree/7dd20894a642a0aa287e9827cb1a1f7f91386b67}{GLM student release}.}
The GLM LoRA configuration uses
$r=32$, $\alpha=64$, zero dropout, and the same $10^{-5}$ learning rate.
It targets the attention projections
\ensuremath{\mathtt{q\_a\_proj}}, \ensuremath{\mathtt{q\_b\_proj}},
\ensuremath{\mathtt{kv\_a\_proj\_with\_mqa}}, \ensuremath{\mathtt{kv\_b\_proj}}, and
\ensuremath{\mathtt{o\_proj}}. Routed experts, embeddings, and the output
head remain frozen. These targets give 42,063,872 adapter parameters
over the 47 decoder layers. The adaptation rank is separate from the
architecture's query/KV compression ranks.\footnote{Projection names and
shapes follow the \href{https://github.com/huggingface/transformers/blob/main/src/transformers/models/glm4_moe_lite/modeling_glm4_moe_lite.py}{Transformers GLM implementation}.
MoE expert tensors require separate targeting when adapted; see the
\href{https://huggingface.co/docs/peft/en/package_reference/lora}{PEFT LoRA reference}.}
Comparisons remain within each model's adapter configuration.

For GLM OPD, we use the frozen teacher
\ensuremath{\mathtt{zai\text{-}org/GLM\text{-}5.3\text{-}Flash}}, FP8 revision
\ensuremath{\mathtt{690b705278a3}}. Its tokenizer file is byte-identical to
the student's. We score the student's unchanged token IDs and causal
prefix, retain its reasoning and tool history, and return sampled-token
log probabilities in FP32.\footnote{See the pinned
\href{https://huggingface.co/zai-org/GLM-5.3-Flash/tree/690b705278a3a58e538fcb37c2ca8b5f9511213c}{GLM teacher release}.
GLM-4.7 and GLM-4.5-Air lack 3,491 student vocabulary entries.}
The teacher's tensors require about 328.25\,GB before caches and runtime
buffers. GLM OPD uses tensor parallelism over two B200 GPUs; the H200
comparison uses Qwen. Before timing, we checked capacity with the
131,072-token context, scoring buffers, and resident quantization state.

Our shared temperature of one differs from the GLM model card's
SWE-bench/Terminal Bench setting of 0.7 at top-$p=1$.
Table~\ref{tab:generation-settings} fixes the other generation limits;
the comparison does not reproduce the model card's evaluation budget.

\subsection{Comparison controls}
Comparisons fix the objective, model, adaptation, initial checkpoint,
training list, rewards, optimizer, precision, serving settings, environment
resources, and hardware across schedules. GRPO and OPD differ in trajectory
counts and supervision costs, so equal updates do not imply equal compute.
Sync and \sysname use one policy snapshot for rollout and learning.
Async follows a \emph{one-step-off} schedule: after startup, each training
batch is sampled from a policy one optimizer update behind the current
learner. Specifically, trajectories are sampled from $\pi_{\theta_{k-1}}$,
while the learner evaluates the same surrogate at $\theta_k$, without an
additional behavior-policy importance correction.
Holding the surrogate fixed makes this a controlled comparison of
execution schedules rather than of staleness corrections such as the
decoupled objective of AReaL~\citep{fu2025areal}.
Fully Async follows the fully asynchronous trainers of
verl~\citep{sheng2025hybridflow} for GRPO and SkyRL~\citep{cao2025skyrl}
for OPD, in the style of AReaL~\citep{fu2025areal}. Rollout and learner
GPUs run independently: the learner starts an update once the completed
groups for one logical batch are available, while rollout continues.
After each update, the new weights reach the rollout engines in flight:
generation pauses, the weights are swapped, and unfinished trajectories
resume under the new policy. A trajectory can therefore span several
policy versions, and one batch can mix trajectories of different ages. A
staleness bound limits how far rollout runs ahead of the learner, so the
policy lag varies but stays bounded.

For Sync, Async, and \sysname, completion order affects admission, not
batch membership: every scheduled trajectory and scored turn completes
before the update. The shared-prefix
learner optimization in Appendix~\ref{app:shared-prefix} also applies to
all schedules. Equal mathematical updates need not be bitwise identical
across kernels.

\section{Evaluation setup}
\label{app:evaluation-setup}

\subsection{Benchmarks and success criteria}
\paragraph{SWE-bench Verified.}
We use the official SWE-bench evaluation harness
({\small$\mathtt{swebench.harness.run\_evaluation}$}) for all 500 tasks in the pinned Verified test
set. This human-validated subset of SWE-bench contains real GitHub issues
paired with repository states and executable tests~\citep{jimenez2024swebench}.
Each attempt receives the issue statement and the repository at its base
commit. The evaluator applies the agent's proposed changes and the
benchmark's test patch in an isolated environment. Resolution requires
all designated \ensuremath{\mathtt{FAIL\_TO\_PASS}} and \ensuremath{\mathtt{PASS\_TO\_PASS}} tests
to pass. Missing test results, invalid patches, and agent timeouts count
as failures. Gold patches and evaluation-only tests are evaluator inputs excluded
from the agent prompt.\footnote{See the
\href{https://huggingface.co/datasets/SWE-bench/SWE-bench_Verified/tree/78f471bf655a3137b2e8a75af1501690ec009ec3}{pinned Verified release}
and the \href{https://www.swebench.com/SWE-bench/guides/evaluation/}{official evaluation guide}.}

\paragraph{Terminal Bench 4.0.}
We use Harbor, the benchmark's official harness, to evaluate all 66 tasks
in release \ensuremath{\mathtt{v4.0.0}},
including their pinned instructions, environments, resource allocations,
and verifiers. The benchmark tests completion of terminal tasks through
changes to the environment~\citep{merrill2026terminalbench}. Each attempt
starts from the task's initial state, and success follows its official
verifier. Reference solutions and verifier files remain outside the
agent's accessible task files. Tasks requiring GPUs or multiple containers
retain those resources in
the evaluation configuration; they are not silently excluded.

Version 4.0 changes task definitions and resources relative to 3.0 and
sets an eight-hour agent timeout.\footnote{See the
\href{https://www.tbench.ai/news/terminal-bench-4-0}{official release description},
\href{https://github.com/harbor-framework/terminal-bench/releases/tag/v4.0.0}{versioned source},
the \href{https://hub.harborframework.com/datasets/terminal-bench/terminal-bench/4}{66-task inventory},
and the \href{https://huggingface.co/datasets/harborframework/terminal-bench}{official Harbor invocation}.}
The protocol retains those task resources and the eight-hour outer timeout,
while imposing the shared token, turn, and ordinary tool-command limits
in Appendix~\ref{app:generation-settings}. Its scores therefore describe
\sysname's specified agent interface and inference budget. They are not
directly interchangeable with leaderboard results from other interfaces
or budgets. SETA's original experiments used Terminal Bench 2.0; their
scores are not baselines for the version-4.0 comparison here.

\paragraph{Training rewards and evaluation success.}
For the declared SWE-smith workload, the training reward is one only when
all designated repair and regression tests pass, and zero otherwise.
For SETA, the selected training reward follows its partial-progress
definition: $r=p+0.5\,\mathbf{1}[p=1]$, where $p$ is the fraction of
passed verifier tests~\citep{shen2026seta}. The same task-specific reward
is used across schedules; OPD records it for monitoring while its
gradient uses teacher scores. Terminal Bench evaluation uses official
task success rather than SETA's partial reward. Consequently, reward
scales should be compared within a workload.

\subsection{Attempts, aggregation, and uncertainty}
The protocol specifies three independent attempts per task at each
evaluated checkpoint, with settings from
Table~\ref{tab:generation-settings}. If $y_{ij}\in\{0,1\}$ indicates
success on attempt $j$ for task $i$, the percentage is
\begin{equation}
  \operatorname{pass@3}
  = \frac{100}{M}\sum_{i=1}^{M}
    \mathbf{1}\!\left[\sum_{j=1}^{3}y_{ij}>0\right],
  \label{eq:benchmark-pass3}
\end{equation}
where $M=500$ for SWE-bench Verified and $M=66$ for Terminal Bench 4.0.
Each task has equal weight. Attempts restart the environment and agent
history; a later attempt receives no solution or verifier feedback from
an earlier one. Pass@3 is the fraction solved at least once in three
attempts. Averaging attempt success estimates pass@1. No fourth attempt is selected
using verifier feedback.

We evaluated training seeds 42, 43, and 44 separately, then averaged their
scores. Three evaluation attempts from one
checkpoint do not constitute three independent training runs.

An infrastructure failure is recorded separately from an agent failure.
A retry is allowed only to replace an attempt invalidated by an
infrastructure fault, using the same task, checkpoint, and assigned seed;
all retries remain in the execution record. A persistently unavailable
task leaves the benchmark evaluation incomplete. It is not dropped from
the denominator to improve the reported score. Agent mistakes, refusals,
and exhausted inference budgets remain failed attempts.

\subsection{Checkpoints and GPU-hours comparisons}
\paragraph{Matched evaluation and training budgets.}
Within each objective and model configuration, \sysname, Sync, Async, and Fully Async use
the same evaluation tasks and settings. Terminal Bench tasks are excluded
from post-training. We report held-out pass@3 and compare schedules by
the time needed to reach the same benchmark-specific pass@3 target and
by pass@3 after the same 128-GPU-hour training budget.
Time to matched performance includes the learning progress across updates;
it is distinct from the per-step speedup defined in
Appendix~\ref{app:efficiency-metrics}.

We requested checkpoint snapshots at one-hour training-time intervals and
saved them at the next completed optimizer boundary, recording their
publication times. The fixed-budget
comparison uses the latest completed checkpoint published no later than
128 GPU-hours. Time to a target is the first evaluated checkpoint attaining
that target, with an unreached target reported as unreached within the
budget.
We fixed the targets before inspecting the comparative runs.

For comparisons across seeds, we formed a common time axis from the union of
checkpoint publication times. At each time, each run contributes its
latest published, evaluated checkpoint score; no future checkpoint is used. The mean of the three runs defines the displayed
step curve. Its first target crossing and its value at 128 GPU-hours use this
same rule.

The training clock starts with the first training rollout request, after
model and environment initialization, whose time is reported separately.
GPU-hours multiply this clock by the eight GPUs of the node.
Training cost in US dollars multiplies GPU-hours by Lambda's on-demand price
for an eight-GPU B200 instance, \$6.69 per GPU-hour (\$53.52 per node-hour,
excluding tax).
It covers rollout, environment interaction, verification,
teacher scoring, learner computation, and policy publication, counting
overlap once. Teacher scoring contributes to elapsed time but its FLOPs
are excluded from the student MFU numerator. Offline benchmark evaluation is timed separately from training.
Checkpointing pauses remain in the training clock. Warmup time excluded
from steady-state step profiling remains in
end-to-end learning curves.

Hyperparameter or prompt changes are selected using development tasks,
then frozen for all schedules. Benchmark checkpoints are evaluated on the
declared cadence rather than selected retrospectively for a favorable
score. Alongside GPU-hours curves, run records must retain optimizer
updates, sampled trajectories, and generated tokens, so that a fixed-time
comparison can be distinguished from a fixed-work comparison.

\paragraph{Checkpoint records.}
Each checkpoint record binds the training seed, update number,
policy hash, publication time, processed trajectories/tokens, and one
success/failure record for each task and attempt. It also stores benchmark,
prompt, tokenizer, harness, and container revisions, plus infrastructure
retry records. We used these records to compute pass@3 and select checkpoints.

\section{Scalability theory}
\label{app:theory}

\subsection{Step time and the learner tail}
\label{app:scalability-timing}

We compare schedules for one logical batch generated and differentiated at
the fixed policy snapshot $\theta_k$. Let $R$ be the time until
the last rollout event required by the objective is available. Learner
job $j$ becomes ready at time $\tau_j$ and takes $p_j$ units of service on
the learner. The total learner service is $L=\sum_{j=1}^{n}p_j$.
After rollout and learning finish, a common serial suffix $P$ covers
final reduction, the optimizer step, policy publication, reload, and
rollout-replica acknowledgement.

The comparison fixes the physical jobs and their service times, rollout
time, hardware allocation, and suffix across schedules. It assumes enough
device memory to admit each source when it becomes ready. In particular,
service times do not change with concurrent rollout activity or with the
schedule's choice of release times. The release laws below assume
$0\leq \tau_j\leq R$; work reserved for the final optimizer barrier belongs
to $P$. These assumptions isolate savings from earlier release.

Index jobs in nondecreasing readiness order. On one learner lane, the
completion time $c_j$ follows
\begin{equation}
    c_0=0,\qquad c_j=\max(\tau_j, c_{j-1})+p_j,
    \qquad T_{\mathrm{\sysnameraw}}=\max(R, c_n)+P.
    \label{eq:scalability-recurrence}
\end{equation}
Repeated substitution gives
\begin{equation}
    c_n=\max_{1\leq j\leq n}
    \left(\tau_j+\sum_{i=j}^{n}p_i\right).
    \label{eq:scalability-unrolled}
\end{equation}
For the induction step, substitute the expression for $c_{j-1}$ into
$\max(\tau_j, c_{j-1})+p_j$: the new job either starts at its own readiness
time or extends one of the earlier uninterrupted service intervals.
Releasing a job earlier keeps or reduces the endpoint when service times and
job order stay fixed. Readiness order also attains the minimum endpoint
on one lane: all jobs released at or after $\tau_j$ must be processed after
that time, giving the lower bound attained in
Equation~\ref{eq:scalability-unrolled}.

Define the learner tail in units of rollout time as
$\lambda_K=[c_n-R]_+/R$, where $[x]_+=\max(x,0)$. Thus $R\lambda_K$ is
the learner time remaining after rollout closes. The two other normalized
quantities are $\rho=L/R$, the learner service relative to rollout time,
and $\kappa=P/R$, the final suffix relative to rollout time. Sync starts
all learner jobs after $R$, so
\begin{equation}
\begin{aligned}
    T_{\mathrm{Sync}}&=R+L+P, &
    T_{\mathrm{\sysnameraw}}&=R(1+\lambda_K)+P,\\
    S_{\mathrm{Sync}}&=\frac{T_{\mathrm{Sync}}}{T_{\mathrm{\sysnameraw}}}
    =\frac{1+\rho+\kappa}{1+\lambda_K+\kappa}.
\end{aligned}
    \label{eq:scalability-speedup}
\end{equation}
Both stages must finish, so every overlap schedule takes at least
$\max(R, L)+P$. Thus,
\begin{equation}
    S_{\mathrm{Sync}}
    \leq\frac{R+L+P}{\max(R, L)+P}\leq 2.
    \label{eq:scalability-ceiling}
\end{equation}
Reaching $2\times$ requires equal stage times, a negligible suffix, and
no learner tail. Equal stage times maximize the ideal overlap opportunity;
the actual peak can shift when a nonzero tail remains.

\subsection{How work release determines the tail}
\label{app:scalability-release-laws}

Write $u=t/R$ for time as a fraction of the rollout window. Let
$F^-(u)$ be the fraction of learner service released strictly before
$uR$:
\begin{equation}
    F^-(u)=\frac{1}{L}\sum_{\tau_j<uR}p_j.
    \label{eq:scalability-release-cdf}
\end{equation}
This fraction weights each job by its service time. Counting ready jobs
alone is insufficient when late trajectories also require more learning.
At a release time, $L[1-F^-(u)]$ includes the newly released jobs and all
later jobs. Equation~\ref{eq:scalability-unrolled} therefore gives
\begin{equation}
    1+\lambda_K
    =\max\left\{1,\sup_{0\leq u\leq1}
        \left[u+\rho\bigl(1-F^-(u)\bigr)\right]\right\}.
    \label{eq:scalability-release-endpoint}
\end{equation}
The strict-before convention includes the service of jobs released at
the same instant. Replacing it with an inclusive fraction would omit
those jobs at their release time.

The maximum lag behind uniform service release is
$D=\sup_{0\leq u\leq1}[u-F^-(u)]$; any trace satisfies
\begin{equation}
    [\rho-1]_+\leq\lambda_K\leq[\rho-1]_++\rho D,
    \qquad \lambda_K=D\quad\text{when }\rho=1.
    \label{eq:scalability-lag-bound}
\end{equation}
To obtain the upper bound, write the expression inside the supremum in
Equation~\ref{eq:scalability-release-endpoint}, minus one, as
$(\rho-1)(1-u)+\rho[u-F^-(u)]$. Its first term is at most
$[\rho-1]_+$ and its second at most $\rho D$. The lower bound follows
from the need to process all $L$ service. At balance, the first term
vanishes, so the tail equals the release lag.

\paragraph{GRPO.}
The plotted model assumes independent trajectory completion times
$U\sim\operatorname{Uniform}(0,1)$ and conditional mean learner service
proportional to $U^{\gamma}$, with $\gamma\geq0$. The fraction of service associated
with trajectories completed by $u$ is then
\begin{equation}
    H(u)=\frac{\int_0^u t^{\gamma}\,dt}{\int_0^1 t^{\gamma}\,dt}=u^{\gamma+1}.
\end{equation}
A fraction $1-\beta$ of service is ready at each trajectory's reward;
the rest, $\beta$, waits for all $m$ trajectories in the group.
By independence, all other members finish with probability $u^{m-1}$, giving
\begin{equation}
    F_{\mathrm{GRPO}}(u)
    =(1-\beta)u^{\gamma+1}+\beta u^{\gamma+m}.
    \label{eq:scalability-grpo-release}
\end{equation}
Substituting this continuous release model into
Equation~\ref{eq:scalability-release-endpoint} gives the plotted tail.
If $\rho[\gamma+1+\beta(m-1)]\leq1$, it is zero. Otherwise an interior
maximum, when present, satisfies
\begin{equation}
    (1-\beta)(\gamma+1)u^{\gamma}+\beta(\gamma+m)u^{\gamma+m-1}=1/\rho.
\end{equation}
The endpoint values cover the boundary cases. Turn count does not appear
after $\rho$, $m$, $\beta$, and the normalized completion and service
distribution are fixed. This explains the flat GRPO surface under these
conditions; changing trajectory length in a real workload can change
those inputs.

\paragraph{OPD.}
For a trajectory completed at $U$, model $a$ equal-service turns with
teacher-scored releases at $jU/a$, $j=1, \ldots, a$. Let $\delta$ be the
fraction of learner service deferred to trajectory completion. Block $j$
is ready by $u$ when $U\leq au/j$. Weighting by the same service model
$U^{\gamma}$ gives
\begin{equation}
    F_{\mathrm{OPD}}(u)
    =\frac{1-\delta}{a}\sum_{j=1}^{a}
      \min\left(1,\frac{au}{j}\right)^{\gamma+1}
      +\delta u^{\gamma+1}.
    \label{eq:scalability-opd-release}
\end{equation}
Each turn can release work before its trajectory closes, including for
late trajectories with large learner service. When $\delta=1$, all work
waits for trajectory completion. For $\delta=0$ and $a\to\infty$, the
release fraction tends to $u(1+1/\gamma)-u^{\gamma+1}/\gamma$ for $\gamma>0$, and to
$u(1-\log u)$ for $\gamma=0$, with value zero at $u=0$. These functions
are at least $u$ on $[0,1]$, so $D=0$ and the tail reaches its minimum
$[\rho-1]_+$. For finite horizons, deferred work and the actual spacing
of teacher scores determine how much of this benefit is realized.

\subsection{Batch size and successive rollout waves}
\label{app:scalability-waves}

Let one full rollout wave take time $W$, with learner service
$\rho_{\mathrm{w}}W$ and normalized tail $\lambda_{\mathrm{w}}$. Consider
$w$ identical full waves launched successively at intervals of $W$,
followed by one final suffix $P$; write $\kappa_{\mathrm{w}}=P/W$.
A job released at normalized time $u$ within wave $i$ has all later
waves' service behind it. Its candidate completion time, divided by $W$,
is
\begin{equation}
    (i-1)+u+\rho_{\mathrm{w}}[1-F^-(u)]
        +\rho_{\mathrm{w}}(w-i).
\end{equation}
Maximizing over $i$ selects the last wave when
$\rho_{\mathrm{w}}\leq1$ and the first when $\rho_{\mathrm{w}}>1$.
After also including rollout completion, the resulting speedup is
\begin{equation}
    S_{\mathrm{Sync}}(w)
    =\frac{w(1+\rho_{\mathrm{w}})+\kappa_{\mathrm{w}}}
    {w\max(1,\rho_{\mathrm{w}})+\kappa_{\mathrm{w}}
       -[\rho_{\mathrm{w}}-1]_++\lambda_{\mathrm{w}}}.
    \label{eq:scalability-wave-speedup}
\end{equation}
At balance this becomes
$(2w+\kappa_{\mathrm{w}})/(w+\kappa_{\mathrm{w}}+\lambda_{\mathrm{w}})$.
The final tail and suffix remain fixed while the overlapped work grows,
so the ratio approaches two. Away from balance, the limit is
$(1+\rho_{\mathrm{w}})/\max(1,\rho_{\mathrm{w}})$.
These statements assume repeated full waves. Continuous refill and
partially occupied waves require either their actual release trace in
Equation~\ref{eq:scalability-release-endpoint} or an explicit approximation.

\subsection{Finite batches and model limitations}
\label{app:scalability-finite}

Equation~\ref{eq:scalability-release-endpoint} is exact for a finite trace
under the lane assumptions. The plotted surfaces replace that trace by
a continuous release distribution. If two release distributions differ
by at most $\epsilon$ at every time, their normalized endpoints differ
by at most $\rho\epsilon$: their expressions inside the supremum differ
pointwise by at most that amount, and taking a supremum and a maximum
preserves this bound:
\begin{equation}
    |\lambda(F,\rho)-\lambda(G,\rho)|
    \leq\rho\sup_u|F(u)-G(u)|.
    \label{eq:scalability-model-error}
\end{equation}
For example, $n$ equal-service atoms placed at the midpoint quantiles of
a continuous distribution have discrepancy at most $1/(2n)$ and tail
error at most $\rho/(2n)$. Random finite batches need a bound for their
sampling process. At fixed $R$ and $L$, replacing a random release
distribution by its mean is optimistic for completion time: the endpoint is a
maximum of affine functions of the release distribution and hence
convex, so its expected value is at least the endpoint of the mean
distribution.

For statically assigned learner queues, apply the recurrence separately
to each rank and take the latest completion. A pooled service rate can
miss a slow rank. Memory pressure that delays admission, packing that
waits for additional sources, unequal turn costs, teacher delays, or
readiness-dependent learner service must also be reflected in the
trace. If concurrent rollout changes kernel service or methods execute different
physical work, the prediction uses service times measured under those conditions.

\subsection{Figure parameters and comparison with measurements}
\label{app:scalability-figure-inputs}

The workload-scaling experiments were run separately to analyze overlap,
with a trajectory-capacity setting of $Q=128$. Their batch sizes, horizons,
and GPU allocations are independent of the reference learning-curve
configuration in Appendix~\ref{app:configurations}; $Q$ is not the
reference vLLM replica's active-sequence limit.

Figures~\ref{fig:scalability-speedup}b,
\ref{fig:scalability-ratio-fixed-traj}, and
\ref{fig:scalability-ratio-fixed-turns} evaluate
Equations~\ref{eq:scalability-grpo-release},
\ref{eq:scalability-opd-release}, and
\ref{eq:scalability-wave-speedup}. For these analytical sweeps, we fix
rollout capacity $Q=128$
trajectories, GRPO group size $m=8$, closure fraction $\beta=0.02$,
OPD deferred fraction $\delta=0.45$, service exponent $\gamma=0.49$, and
suffix $\kappa_{\mathrm{w}}=0.05$.

Let $B$ denote trajectories per optimizer step and $r$ the rollout time
divided by learner time at full capacity. The plotting approximation uses
\begin{equation}
    w=\max(1,B/Q),\qquad
    \rho_{\mathrm{w}}=\frac{\min(1,B/Q)}{r}.
\end{equation}
Figure~\ref{fig:scalability-speedup}b fixes $r=1$. For $B<Q$, the
model keeps one rollout window and reduces learner service in proportion
to occupancy. For noninteger $B/Q>1$, it extends the full-wave formula
continuously. These approximations extend the full-wave model to partial
occupancy. The ratio sweeps vary $r$; their actual
speedup peak need not be exactly at one.

Figure~\ref{fig:speedup-prediction} compares the analytical surfaces with
measured release traces and wall-clock speedups. We evaluated 588 paired
configurations, 294 per objective. At 40 turns, the batch
grid is $B\in\{64,\allowbreak96,\allowbreak128,\allowbreak192,\allowbreak256,\allowbreak384,\allowbreak512,\allowbreak768,\allowbreak1024,\allowbreak1536,\allowbreak2048\}$;
at $B=512$, the horizon grid is
$a\in\{10,\allowbreak15,\allowbreak20,\allowbreak30,\allowbreak40,\allowbreak60,\allowbreak100,\allowbreak150,\allowbreak200,\allowbreak300\}$. Each grid point uses seven
nominal rollout-to-learner ratios
$\{0.35,0.5,0.7,1,1.4,2,2.8\}$ and two repetitions, giving
$11\times7\times2=154$ and $10\times7\times2=140$ configurations.
The plotted ratio uses the measured stage times for the selected GPU
allocation, rather than the ratio of GPU counts. The horizon sweep varies
the turn limit up to 300; its turn limit and resulting context lengths
are recorded separately from the learning-curve runs.

Each configuration has an analytical prediction $s_i$, a trace prediction
$o_i$ (open marker), and a measured wall-clock speedup $f_i$ (filled marker).
We computed $o_i$ by replaying the observed source readiness and service
times through Equation~\ref{eq:scalability-recurrence}. The ratio $f_i$
compares Sync and \sysname with the same model, logical workload, GPU roles,
precision, and generation settings within the cell. These controls hold
within a configuration even when the allocation changes between cells.

The reported differences are computed for each paired configuration as
\begin{equation}
    e_i=100\frac{|o_i-s_i|}{s_i},\qquad
    d_i=100\frac{o_i-f_i}{o_i}.
\end{equation}
Across both objectives and both rows, the mean of $e_i$ is $1.5911\%$
and its maximum is $10.5541\%$; the mean of $d_i$ is $6.5148\%$.
The main text rounds these to $1.6\%$, $10.6\%$, and $6.5\%$.
The first difference measures the release model's agreement with the
trace prediction; the second measures the remaining gap to runtime
execution. The common optimizer and publication suffix is already present
through $\kappa_{\mathrm{w}}$, so the second difference does not represent
the entire publication cost.

\paragraph{Trace collection and calibration.}
We recorded a manifest for each objective, model, batch,
horizon, and integer GPU allocation, including tensor/data parallelism,
teacher allocation, environment concurrency, and memory limits. Sync and
\sysname executed the same logical workload within each cell. The measured
$R$, $L$, and $P$ then determine the cell's time ratio, so each ratio reflects
measured service rates.

Each trace records a source's task, trajectory, group, policy version,
learner rank, readiness time, service start/end, scored-token count, and
deferred-work completion. GRPO readiness is receipt of the required reward;
OPD readiness is receipt of valid teacher scores. Device events measure
service, while a shared host clock records arrival and publication
acknowledgements. Replay uses both $\tau_j$ and $p_j$ in
Equation~\ref{eq:scalability-recurrence}, with separate queues for statically
assigned ranks. Teacher waiting and deferred backward stay in the trace.
We calibrated release-law parameters on separate traces and held them
fixed for evaluation. Profiling windows follow
Appendix~\ref{app:efficiency-metrics}; missing or infeasible cells are
reported explicitly.

\subsection{Balance sweeps}
\label{app:scalability-grids}

\begin{figure}[h!]
  \centering
  \includegraphics[width=\linewidth]{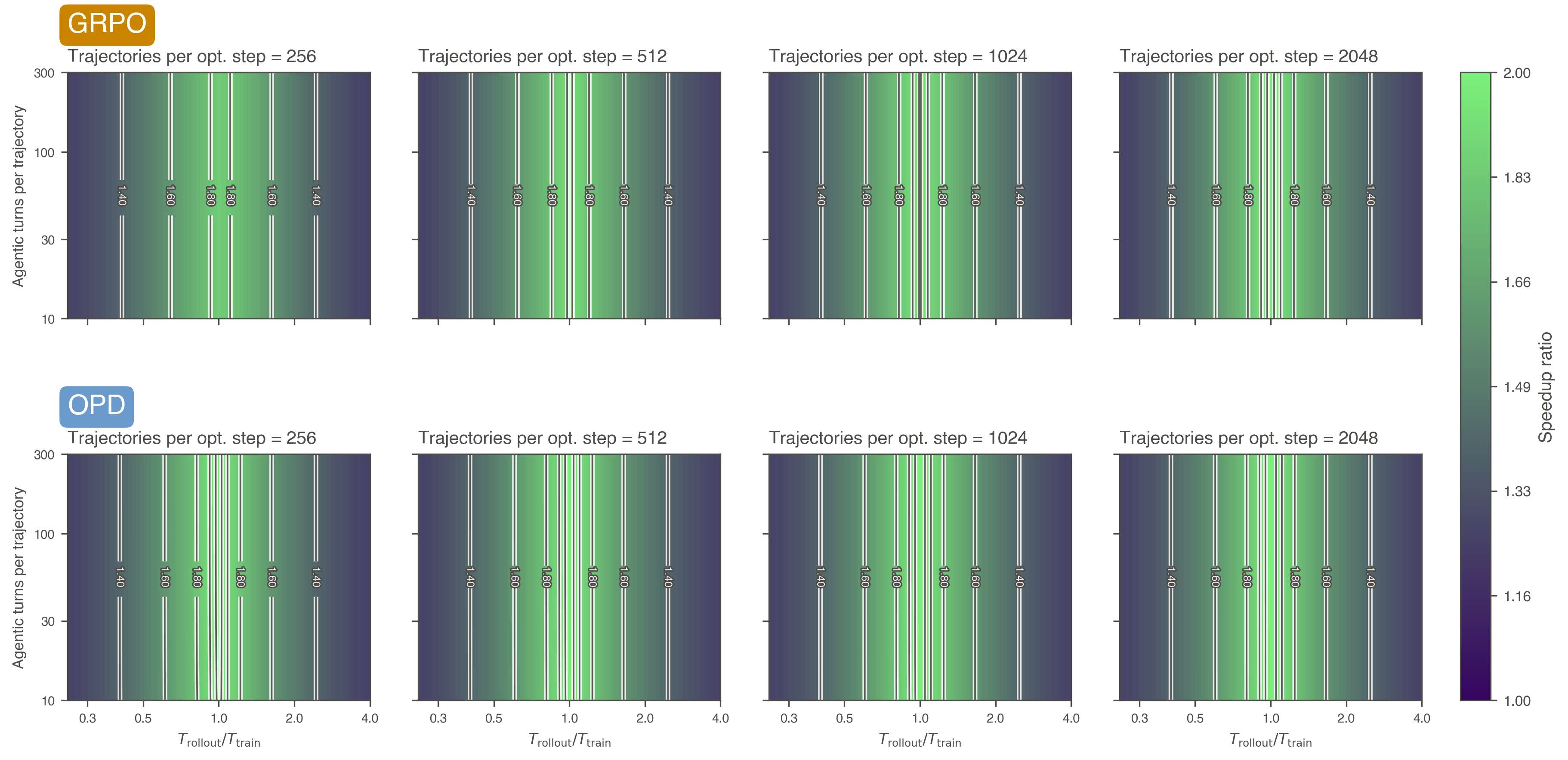}
  \caption{Speedup over Sync as a function of the rollout-to-training time ratio and agentic turns per trajectory, at fixed trajectories per optimizer step (256 to 2048), for GRPO (top) and OPD (bottom).}
  \label{fig:scalability-ratio-fixed-traj}
\end{figure}

\begin{figure}[h!]
  \centering
  \includegraphics[width=\linewidth]{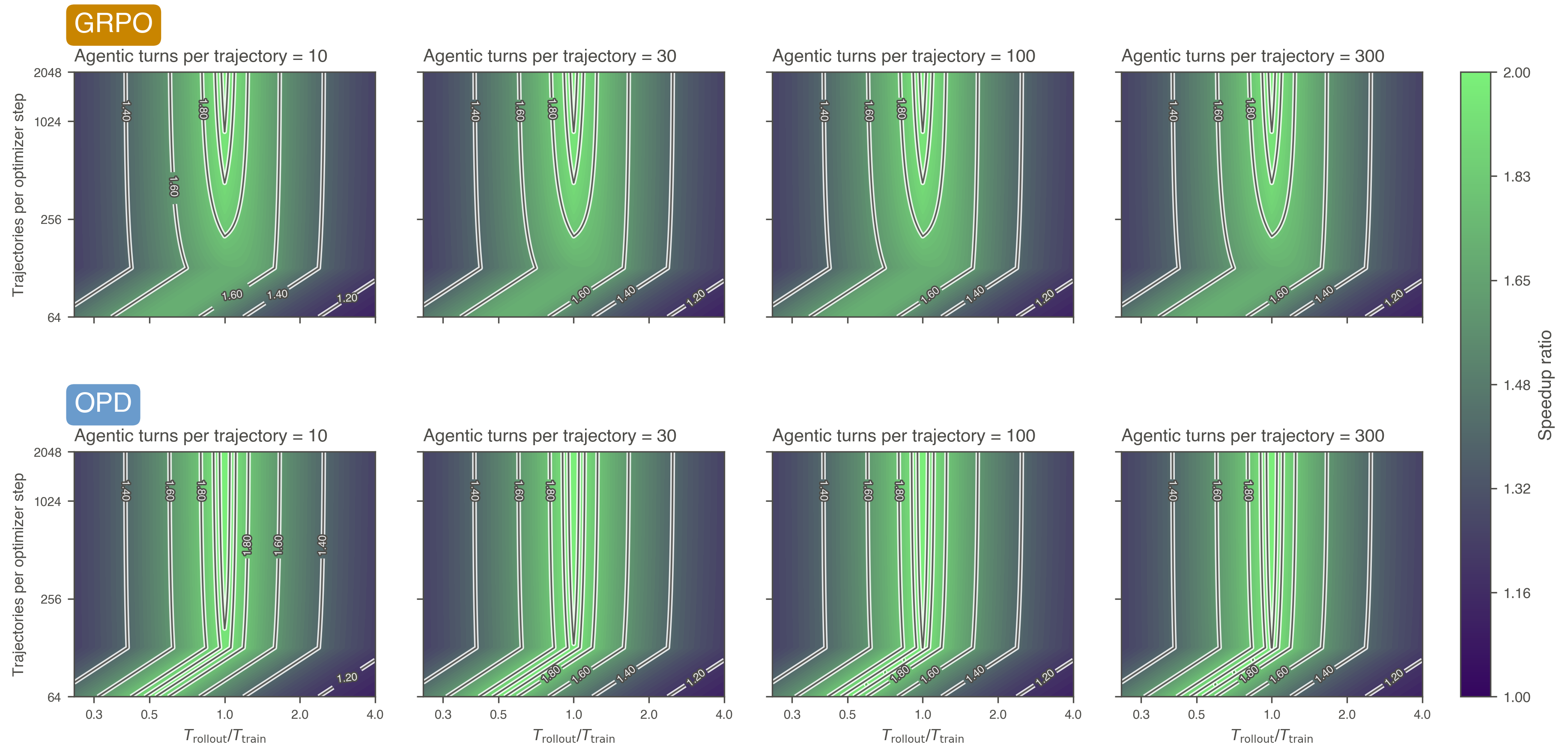}
  \caption{Speedup over Sync as a function of the rollout-to-training time ratio and trajectories per optimizer step, at fixed agentic turns per trajectory (10 to 300), for GRPO (top) and OPD (bottom).}
  \label{fig:scalability-ratio-fixed-turns}
\end{figure}

Figures~\ref{fig:scalability-ratio-fixed-traj}
and~\ref{fig:scalability-ratio-fixed-turns} sweep the rollout-to-learner
ratio $r$ from $0.3$ to $4$ with the parameters of
Appendix~\ref{app:scalability-figure-inputs}. Both figures are analytical
evaluations of Equation~\ref{eq:scalability-wave-speedup}; they contain no
measured points. Figure~\ref{fig:scalability-ratio-fixed-traj} fixes the
batch size and varies the horizon, and
Figure~\ref{fig:scalability-ratio-fixed-turns} fixes the horizon and
varies the batch size.

For batches that fill at least one rollout wave ($B\geq Q$), every panel
peaks near balance, $r\approx1$. There, the predicted speedup is
$1.84\times$ for GRPO at 256 trajectories per step and $1.98\times$ at 2,048;
OPD reaches $1.97$ to $1.98\times$ at 256 and $2.00\times$ at 2,048. Away from
balance, the larger stage bounds the step, and the speedup falls toward the
full-overlap limit $(1+\rho_{\mathrm{w}})/\max(1,\rho_{\mathrm{w}})$ of
Appendix~\ref{app:scalability-waves}, which is symmetric in $\log r$.
For $B\geq Q$, halving or doubling the ratio caps the speedup near
$1.5\times$ ($1.46$ to $1.50\times$), $r=0.35$ or $2.8$ gives $1.34$ to
$1.36\times$, and $r=4$ gives $1.24$ to $1.25\times$, for both objectives.
Below one full wave, the plotting rule scales learner service by $B/Q$, so
balance moves to $r\approx B/Q$: at $B=64$ the peak sits near $r=0.5$, with
$1.71\times$ for GRPO and $1.95\times$ for OPD. This bends the bright band
toward smaller ratios at the bottom of
Figure~\ref{fig:scalability-ratio-fixed-turns}.

The horizon matters only near balance. GRPO's surface does not change with
turn count, as Appendix~\ref{app:scalability-release-laws} explains. For
OPD, more turns spread releases and shorten the tail, but at 256 trajectories
per step the speedup at $r=1$ moves only from $1.967\times$ at 10 turns to
$1.975\times$ at 300 turns. Batch size has the larger effect:
more waves amortize the final tail and suffix, which is why the bright
band at $r\approx1$ widens as trajectories per step grow in
Figure~\ref{fig:scalability-ratio-fixed-turns}.

\section{Empirical update equivalence}
\label{app:update-equivalence}

We verify the equivalence of Section~\ref{sec:gradient-streaming} on the
B200 and H200 nodes of Appendix~\ref{app:configurations}, comparing the
updates of Sync and \sysname on identical logical batches.

\paragraph{Protocol.}
For each objective, we record ten consecutive logical batches of a
Qwen-3.8-27B Sync run, including tokens, scored-token masks, rewards, and
teacher scores, and replay them under Sync and \sysname from the same initial
parameters $\vtheta_0$. \sysname ingests trajectories and scored turns in their
recorded completion order. Replaying fixed batches isolates the learner
computation. Both schedules use the optimizer settings of
Table~\ref{tab:optimization-settings} and run twice: with the BF16 student
forward used in training and with an FP32 forward as the reference. We compare
the first batch direction $\vg_0$ before clipping, coordinate by coordinate,
and after each update $k$ report the parameter gap
\begin{equation}
  \Delta_k =
  \frac{\|\vtheta^{\mathrm{ours}}_k-\vtheta^{\mathrm{sync}}_k\|}
       {\|\vtheta^{\mathrm{sync}}_k-\vtheta_0\|},
  \label{eq:parameter-gap}
\end{equation}
the distance between the two schedules relative to the distance that training
has moved the parameters.

\begin{figure}[ht!]
  \centering
  \includegraphics[width=\linewidth]{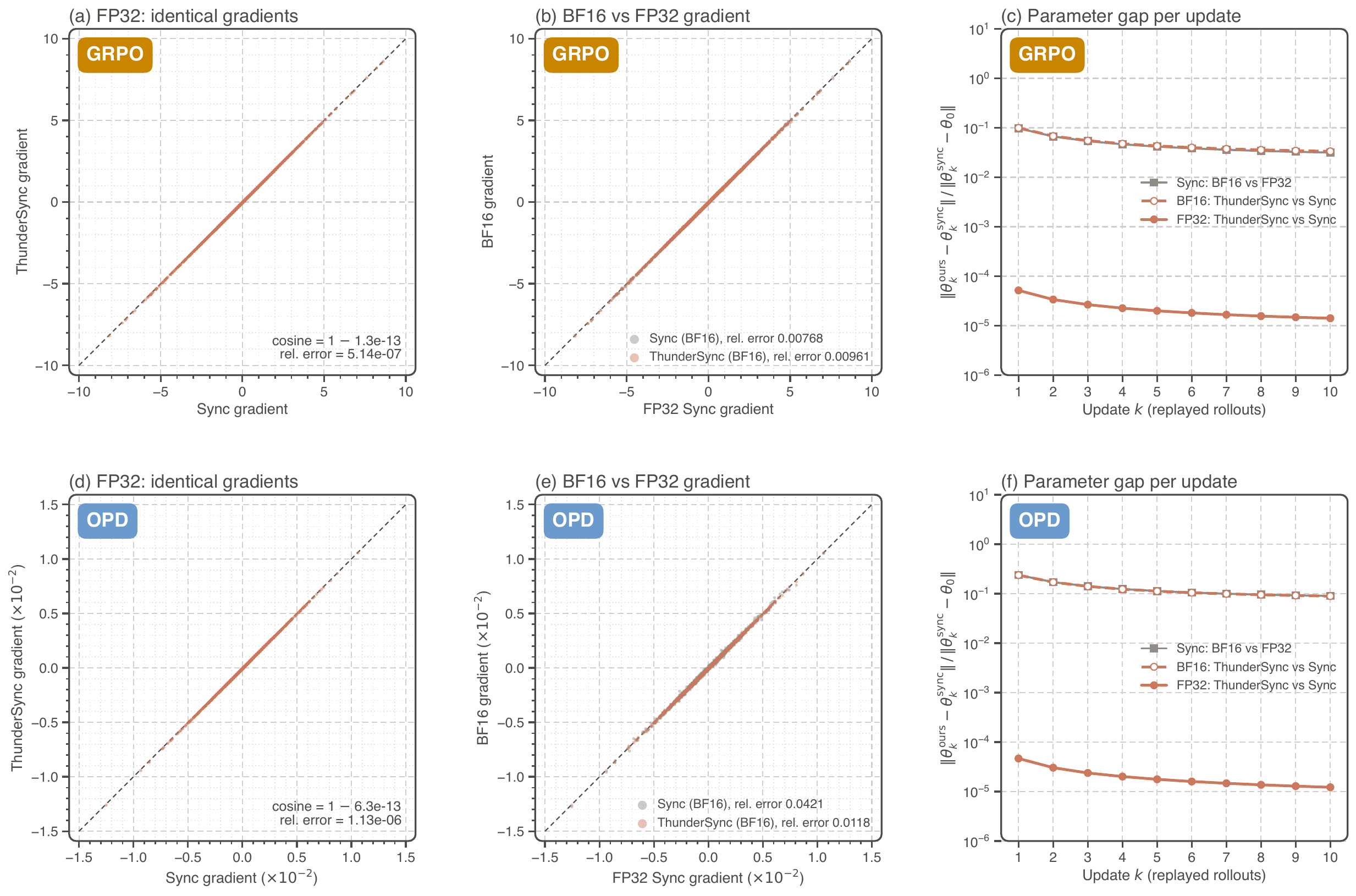}
  \caption{Update equivalence on B200 for GRPO (top row) and OPD (bottom
  row). (a, d)~Each point is one randomly sampled coordinate of the first
  batch direction $\vg_0$ with FP32 arithmetic, \sysname against Sync. (b, e)~Both schedules
  with BF16 arithmetic against the FP32 Sync direction. (c, f)~Parameter gap
  $\Delta_k$ (Equation~\ref{eq:parameter-gap}) after each of ten replayed
  updates, with the gap between BF16 and FP32 Sync as reference. Dashed diagonals
  mark equality.}
  \label{fig:update-parity-b200}
\end{figure}

\begin{figure}[ht!]
  \centering
  \includegraphics[width=\linewidth]{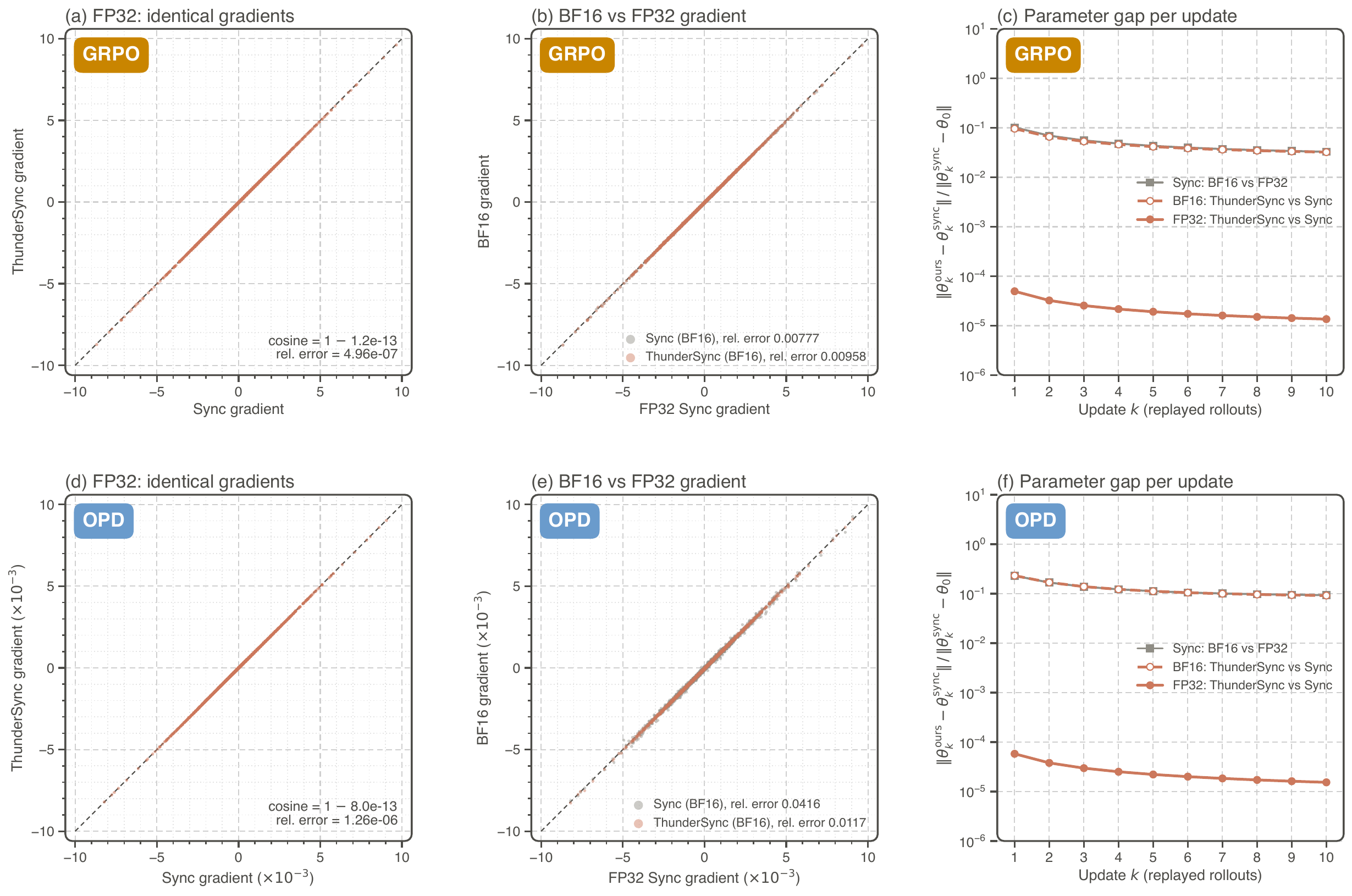}
  \caption{Update equivalence on H200, with panels as in
  Figure~\ref{fig:update-parity-b200}.}
  \label{fig:update-parity-h200}
\end{figure}

\paragraph{Results.}
Figures~\ref{fig:update-parity-b200} and~\ref{fig:update-parity-h200} show
the same result on both GPU types. In FP32, the \sysname and Sync gradients
coincide coordinate by coordinate (panels a and d), with relative errors of
$5.1\times10^{-7}$ for GRPO and $1.1\times10^{-6}$ for OPD on B200. In BF16,
both schedules stay within a few percent of the FP32 gradient (panels b and e):
relative errors are $0.77\%$ for Sync and $0.96\%$ for \sysname under GRPO,
and $4.2\%$ and $1.2\%$ under OPD, where Sync's BF16 error is larger.
Across ten updates, the FP32 parameter gap stays below $6\times10^{-5}$,
nearly three orders of magnitude below the gap between BF16 and FP32 Sync, and the
BF16 gap between the schedules follows that same BF16 curve (panels c and f).
\sysname reproduces the batch-synchronous update at every step within the
numerical precision of training.

\paragraph{Binary-reward clipped-PPO extension.}
We extend the replay comparison to the clipped PPO surrogate with binary
rewards, including subsequent minibatch updates and repeated optimization
epochs. Both schedules use identical rollout data, behavior
log-probabilities, minibatch order, and initial model and optimizer states.
Gradients are recomputed at each update, and the comparison includes later
updates with active surrogate clipping. We compare gradient directions
before global norm clipping and parameter gaps after each optimizer step,
following the definitions above.
In the FP32 replay, the maximum gradient relative error is
$2.1\times10^{-6}$ and the maximum normalized parameter gap is
$6.4\times10^{-5}$. If confirmed experimentally, these results would support
update equivalence within numerical precision for the tested binary-reward
clipped-PPO extension, including updates where the learner differs from
the behavior policy.

\clearpage
\section{Further Experimental Results}
\label{app:further-experiments}

\subsection{Pass@1 and Pass@5}
\label{app:pass-at-1-and-5}

Pass@$k$ is the fraction of tasks solved at least once in $k$ independent
attempts. Following \citet{chen2021codex}, we estimate it from $n \ge k$
attempts per task: if $c_i$ of the $n$ attempts on task $i$ succeed,
\begin{equation}
  \operatorname{pass@}k
  = \frac{100}{M}\sum_{i=1}^{M}
    \left(1 - \binom{n-c_i}{k}\Big/\binom{n}{k}\right),
  \label{eq:pass-at-k}
\end{equation}
which is unbiased for the probability that at least one of $k$ attempts
succeeds. Pass@1 is the mean attempt success rate. With $n=k$ the estimator
reduces to the empirical rate of tasks solved at least once, the form of
pass@3 in Equation~\ref{eq:benchmark-pass3}.

\begin{figure}[ht!]
  \centering
  \includegraphics[width=\linewidth]{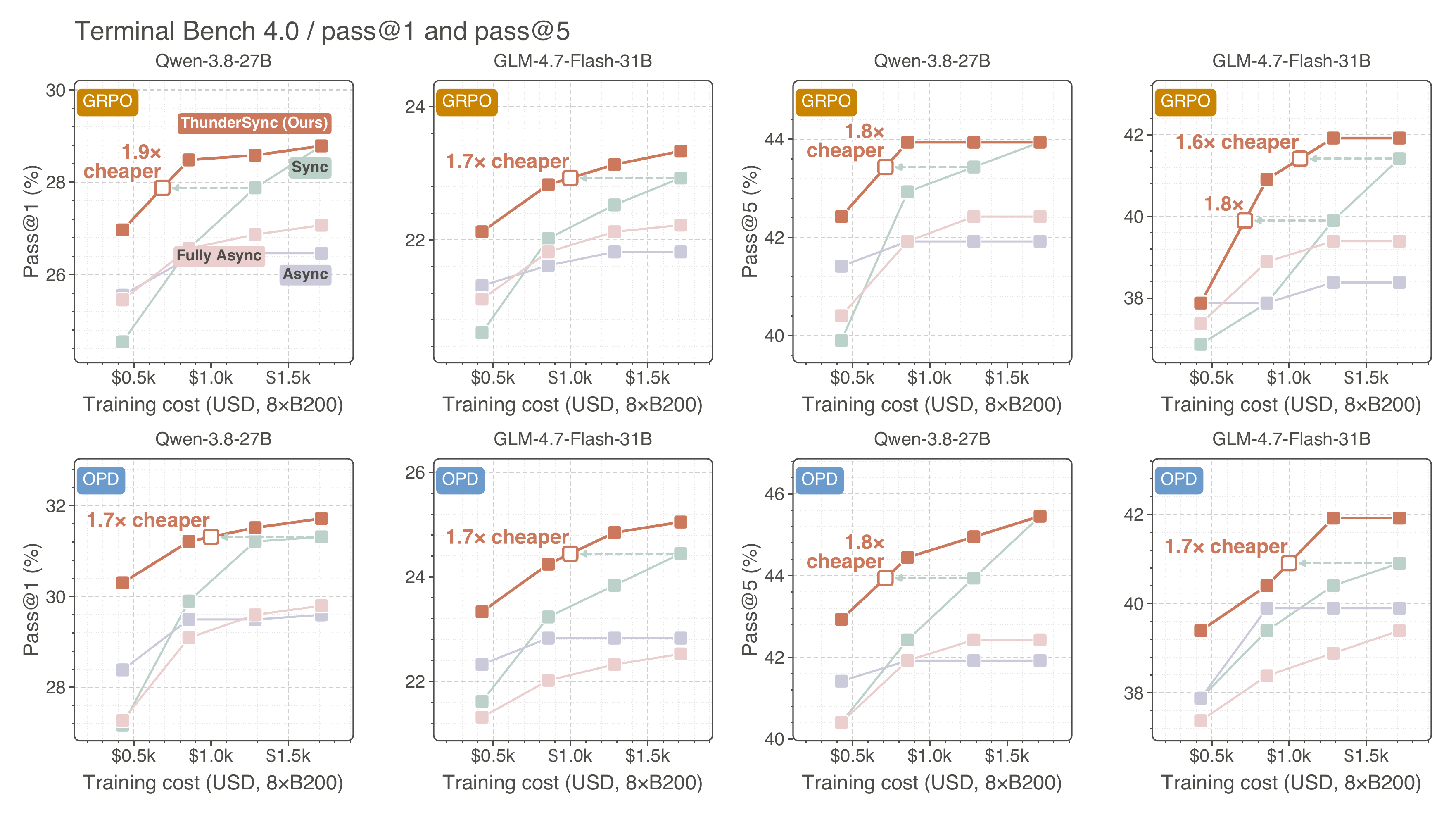}
  \caption{Pass@1 (left two panels) and pass@5 (right two panels) on Terminal Bench 4.0 against training cost for Qwen-3.8-27B and GLM-4.7-Flash-31B under GRPO (top row) and OPD (bottom row). Arrows mark where \sysname reaches the score of a Sync checkpoint; labels give the cost ratio.}
  \label{fig:pass-at-1-and-5-terminal-bench}
\end{figure}

For this appendix, each evaluated checkpoint runs five independent attempts
per task ($n=5$), from which both pass@1 and pass@5 are computed; the other
settings follow Table~\ref{tab:generation-settings}, and attempts restart the
environment and agent history as in Appendix~\ref{app:evaluation-setup}. Figures~\ref{fig:pass-at-1-and-5-swebench}
and~\ref{fig:pass-at-1-and-5-terminal-bench} report pass@1 and pass@5 against
training cost for Qwen-3.8-27B and GLM-4.7-Flash-31B under GRPO and OPD, in
the form of Figure~\ref{fig:headline}. Pass@1 scores a single attempt;
pass@5 credits a task solved in any of the five.

\begin{figure}[ht!]
  \centering
  \includegraphics[width=\linewidth]{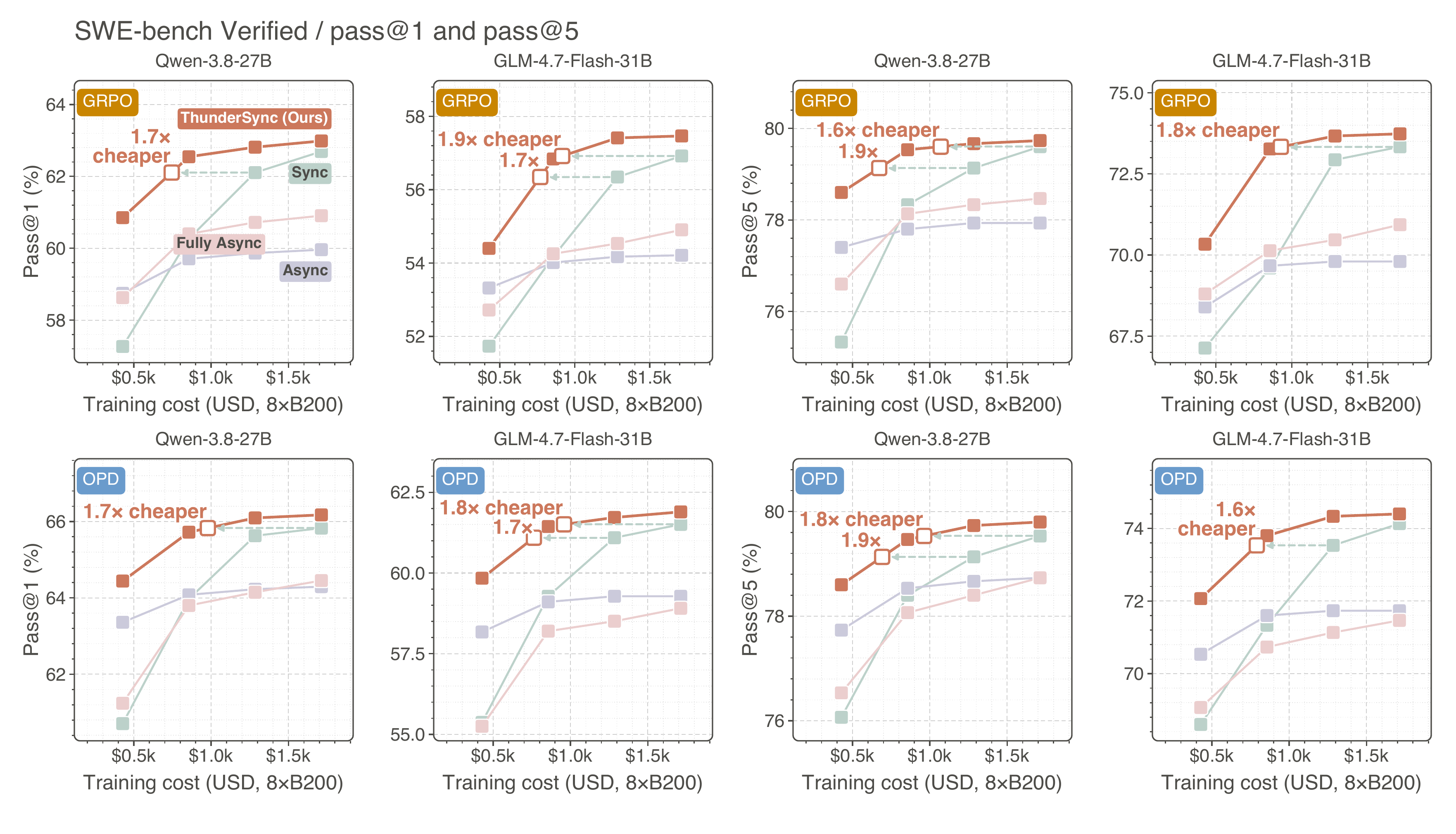}
  \caption{Pass@1 (left two panels) and pass@5 (right two panels) on SWE-bench Verified against training cost for Qwen-3.8-27B and GLM-4.7-Flash-31B under GRPO (top row) and OPD (bottom row). Arrows mark where \sysname reaches the score of a Sync checkpoint; labels give the cost ratio.}
  \label{fig:pass-at-1-and-5-swebench}
\end{figure}

\subsection{Efficiency with LoRA}
\label{app:lora-efficiency}

We profiled the Qwen and GLM LoRA configurations on eight B200 GPUs
(Table~\ref{tab:lora-efficiency}). Each configuration uses four learner GPUs,
replicated BF16 frozen weights, FP32 adapter state, and the objective-specific
rollout and teacher allocation in Appendix~\ref{app:configurations}.

\begin{table}[ht!]
  \centering
  \caption{LoRA efficiency on NVIDIA B200 GPUs.
  Each speedup uses the same model and objective's Sync reference.
  OPD model names refer to the student.}
  \label{tab:lora-efficiency}
  \begingroup
  \small
  \setlength{\tabcolsep}{2pt}
  \renewcommand{\arraystretch}{1.00}
  \resizebox{\linewidth}{!}{%
  \begin{tabular}{@{}lcccc@{\hspace{14pt}}ccc@{}}
    \toprule
    \multirow{2}{*}{Method} &
    \multirow{2}{*}{\shortstack{Policy lag\\(staleness) $\downarrow$}} &
    \multicolumn{3}{c}{\textbf{GRPO}} &
    \multicolumn{3}{c}{\textbf{OPD}} \\
    \cmidrule(lr){3-5}\cmidrule(l){6-8}
    & & \shortstack{Trainer active\\MFU (\%) $\uparrow$} &
    \shortstack{Speedup\\($\times$) $\uparrow$} &
    \shortstack{Peak GPU\\memory (GB) $\downarrow$} &
    \shortstack{Trainer active\\MFU (\%) $\uparrow$} &
    \shortstack{Speedup\\($\times$) $\uparrow$} &
    \shortstack{Peak GPU\\memory (GB) $\downarrow$} \\
    \midrule
    \multicolumn{8}{@{}l}{\textbf{Qwen-3.8-27B}\quad LoRA} \\
    \addlinespace[2pt]
    \methoddot{methodSync}Sync & 0 & 19.5 & 1.00 & 75.4 & 19.0 & 1.00 & 73.4 \\
    \methoddot{methodAsync}Async & 1 & 32.6 & 1.67 & 76.0 & 32.1 & 1.69 & 74.0 \\
    \methoddot{methodFullyAsync}Fully Async & Variable & 32.8 & 1.68 & 77.1 & 32.3 & 1.70 & 74.8 \\
    \methoddot{methodThunderSync}\textbf{\sysname} & 0 & 33.0 & 1.69 & 79.2 & 32.5 & 1.71 & 76.0 \\
    \midrule
    \multicolumn{8}{@{}l}{\textbf{GLM-4.7-Flash-31B}\quad LoRA} \\
    \addlinespace[2pt]
    \methoddot{methodSync}Sync & 0 & 10.9 & 1.00 & 75.1 & 10.3 & 1.00 & 73.1 \\
    \methoddot{methodAsync}Async & 1 & 17.7 & 1.62 & 75.6 & 16.5 & 1.60 & 73.6 \\
    \methoddot{methodFullyAsync}Fully Async & Variable & 17.8 & 1.63 & 76.3 & 16.6 & 1.61 & 74.2 \\
    \methoddot{methodThunderSync}\textbf{\sysname} & 0 & 17.9 & 1.64 & 77.3 & 16.7 & 1.62 & 75.0 \\
    \bottomrule
  \end{tabular}%
  }
  \endgroup
\end{table}

For Sync, Async, and \sysname, LoRA retains the efficiency ordering of full fine-tuning.
\sysname reaches 16.7\% to 33.0\% trainer active MFU, slightly above Async's
16.5\% to 32.6\% and above Sync's 10.3\% to 19.5\%.
Its step wall time speedups are $1.69/1.71\times$ for Qwen and
$1.64/1.62\times$ for GLM under GRPO/OPD, compared with Async's
$1.67/1.69\times$ and $1.62/1.60\times$, respectively.
\sysname retains zero policy lag in every configuration.

For Qwen, the speedups are smaller than under full fine-tuning; for GLM,
whose step is already bound by rollout, they barely change. Freezing base weights
removes their parameter-gradient and optimizer work, reducing the learner
work available to overlap with rollout and teacher scoring. Forward and
input-gradient computation through those weights remains necessary.
Trainer active MFU counts learner FLOPs over the complete update wall
time, including rollout waiting. It therefore reflects both the reduced
learner work under LoRA and the idle time remaining after overlap.

For these three schedules, trainer GPU memory is lower with LoRA than with full fine-tuning.
\sysname's GPU peaks range from 75.0 to 79.2\,GB per learner GPU, with an additional
1.9 to 3.8\,GB relative to Sync for live gradient statistics and deferred
state. The measurements include frozen weights, adapter and optimizer
state, activations, and temporary buffers; the separate rollout and
teacher pools are excluded. The GLM teacher and adapter targets are
specified in Appendix~\ref{app:glm-settings}.

\section{\sysname algorithm and implementation}
\label{app:implementation}

Algorithm~\ref{alg:thundersync} summarizes one \sysname update. The rollout and learner
threads run concurrently at the fixed policy snapshot $\vtheta_k$.
For GRPO, a ready source contains a trajectory's scored action tokens and
reward. For OPD, it contains a turn's scored action tokens and frozen-teacher
scores. Completion events release deferred context backward and group
reduction once their dependencies are satisfied. The implementations below
give the objective-specific normalization and optimizer conventions.

\begingroup
\algrenewcommand{\algorithmiccomment}[1]{\hfill\textit{(#1)}}
\begin{algorithm}[H]
\caption{\sysname: gradient streaming with a synchronous update}
\label{alg:thundersync}
\small
\begin{algorithmic}[1]
    \State \textbf{Input:} Policy $\pi_{\vtheta_k}$, batch $\mathcal{B}_k$, objective $o$, frozen teacher $\pi_{\mathrm{T}}$ for OPD
    \State \textbf{Output:} Updated $\vtheta_{k+1}$, acknowledged by every rollout replica
    \State $\mathcal{Q}\gets\operatorname{Queue}()$; register all trajectories and groups in $\mathcal{B}_k$
    \State Initialize objective accumulators $\mathcal{A}$; keep $\vtheta_k$ fixed
    \State
    \State \textbf{Thread 1: Rollout and supervision (producer)}
    \ForAll{trajectories $i\in\mathcal{B}_k$ \textbf{in parallel}}
        \State Generate with $\pi_{\vtheta_k}$; enqueue ready sources and completion events as they occur
    \EndFor
    \State
    \State \textbf{Thread 2: Learner (consumer)}
    \While{a registered trajectory or group remains unfinished}
        \State $e\gets\mathcal{Q}.\operatorname{pop}()$ \Comment{Wait for the next available event}
        \If{$e$ is a GRPO trajectory source $i$ in group $\mathcal{G}$}
            \State $\mH_i\gets\left.\nabla_{\vtheta}S_i(\vtheta)\right|_{\vtheta=\vtheta_k}$
            \State $\mG^{(1)}_{\mathcal{G}}\gets\mG^{(1)}_{\mathcal{G}}+r_i\mH_i$; $\mG^{(2)}_{\mathcal{G}}\gets\mG^{(2)}_{\mathcal{G}}+\mH_i$
        \ElsIf{$e$ is an OPD turn source $j$}
            \State Accumulate $\vg_j$ and $|\mathcal{M}_j|$ in $\mathcal{A}$ \Comment{Equation~\ref{eq:opd-source-gradient}}
        \EndIf
        \State Update completion state; finish eligible group reductions and deferred backward
    \EndWhile
    \State
    \State \textbf{Synchronous batch barrier}
    \State Wait for all rollouts, source gradients, and deferred work in $\mathcal{B}_k$ to finish
    \State $\widehat{\vg}_k\gets\operatorname{DistributedReduceAndNormalize}(\mathcal{A},o)$
    \State $\vtheta_{k+1}\gets\operatorname{Opt}\!\left(\vtheta_k,\operatorname{clip}(\widehat{\vg}_k)\right)$ \Comment{Exactly one optimizer step}
    \State Publish $\vtheta_{k+1}$; wait for every replica to reload, clear caches, and acknowledge
    \State \Return $\vtheta_{k+1}$
\end{algorithmic}
\end{algorithm}
\endgroup

\paragraph{Shared-prefix backward in all schedules.}
\label{app:shared-prefix}
Sync, Async, and \sysname all reuse shared-prefix computation on the learner.
Training-time prefix reuse has earlier applications in preference
optimization~\citep{wang2024prefixsharing} and in branching agent
trajectories~\citep{wang2025treetraining}.
At a fixed learner snapshot, we compute the shared prefix once, process
each continuation with its causal context, and accumulate its gradient
with respect to the prefix boundary states. After all branch contributions
are available, one backward pass through the shared prefix propagates
their sum, with the objective's weights and normalization preserved.
For GRPO, the reward-weighted and unweighted boundary contributions are
combined when the group rewards are fixed. For OPD, teacher-scored turns
contribute boundary gradients before the final token normalization.
This is a common learner optimization in the comparison, distinct from
vLLM's inference-time prefix cache. \sysname admits eligible branch work
earlier; shared-prefix backward remains part of each schedule's measured
learner work and update time.

\ifdefined\arxivversion\else\clearpage\fi

\begingroup
\definecolor{thunderSyncCodeBackground}{RGB}{245,245,245}
\definecolor{thunderSyncCodeRule}{RGB}{165,165,165}
\definecolor{thunderSyncCodeKeyword}{RGB}{235,0,135}
\definecolor{thunderSyncCodeString}{RGB}{170,0,255}
\definecolor{thunderSyncCodeComment}{RGB}{0,145,0}
\renewcommand{\ttdefault}{zi4}
\renewcommand{\lstlistingname}{Implementation}
\lstdefinestyle{thunderSyncImplementation}{
    language=Python,
    basicstyle=\fontsize{7.7}{8.8}\fontencoding{T1}\ttfamily\selectfont,
    keywordstyle=\color{thunderSyncCodeKeyword}\bfseries,
    stringstyle=\color{thunderSyncCodeString},
    commentstyle=\color{thunderSyncCodeComment},
    backgroundcolor=\color{thunderSyncCodeBackground},
    numbers=left,
    numberstyle=\fontsize{6}{8.8}\selectfont\color{gray},
    numbersep=7pt,
    stepnumber=1,
    numberblanklines=true,
    showstringspaces=false,
    keepspaces=true,
    columns=fullflexible,
    tabsize=4,
    breaklines=true,
    breakatwhitespace=true,
    frame=tb,
    rulecolor=\color{thunderSyncCodeRule},
    framerule=0.6pt,
    framesep=4pt,
    xleftmargin=14pt,
    xrightmargin=4pt,
    aboveskip=8pt,
    belowskip=0pt,
    captionpos=b,
    abovecaptionskip=6pt,
    belowcaptionskip=0pt
}
\makeatletter
\long\def\lst@makecaption#1#2{%
    \vskip\lst@abovecaption
    {\small\noindent\textit{#1.} #2\par}%
    \vskip\lst@belowcaption
}
\makeatother

The following Python-style pseudocode describes one logical update of \sysname.
Rollout workers and the learner use the same fixed policy version throughout
the update. Each learner rank processes its assigned groups. Helpers represent
model execution, gradient storage, and distributed collectives.
\ensuremath{\mathtt{publish\_and\_wait}} publishes the updated
parameters and waits for every rollout replica to load them, clear
policy-dependent caches, and acknowledge the new version.

\subsection{GRPO}
\label{app:implementation-grpo}

Implementation~\ref{impl:thundersync-grpo} shows reward-ready admission and the
decomposition in Equation~\ref{eq:thundersync-decomposition}. The source gradient
can also be prepared when content is complete
and bound to its reward later.
Gradient arithmetic includes any captured boundary adjoints;
\ensuremath{\mathtt{finish\_group\_backward}} propagates their combined contribution through
deferred context. The code applies the sum of group directions,
$\vg_k=\sum_{\mathcal{G}}\vg_{\mathcal{G}}$, as in
Section~\ref{sec:gradient-streaming}; the token weights in
Appendix~\ref{app:configurations} make this sum the mean over the
trajectory-normalized directions.

\begin{lstlisting}[style=thunderSyncImplementation,label={impl:thundersync-grpo},caption={\sysname GRPO update. Each trajectory releases its score gradient independently; group closure fixes the reward coefficients. The optimizer uses gradient ascent.}]
def thundersync_grpo_step(model, rollout, batch, optimizer, eps, max_norm):
    """One reward-linear GRPO epoch; optimizer has maximize=True."""
    params = trainable_parameters(model)
    version = rollout.policy_version
    groups = register_groups(batch)  # Sizes, rewards, zero G1/G2 buffers
    batch_grad = zeros_like(params)
    optimizer.zero_grad(set_to_none=True)

    # 1. Start learner work while other trajectories are still running.
    for src in rollout.reward_ready_sources(batch, version):
        assert src.policy_version == version
        group = groups[src.group_id]
        logp = rebuild_scored_logprobs(model, src)  # Fixed action mask
        score = (src.token_weights * logp).sum()
        H = backward_with_boundary_capture(score, params)

        # 2. Fold each completed source once into its group statistics.
        group.G1 += src.reward * H
        group.G2 += H
        group.rewards.append(src.reward)
        group.num_reduced += 1

        # 3. Combine only after every reward and source gradient arrives.
        if group.num_reduced == group.size:
            r = torch.tensor(group.rewards, dtype=torch.float64)
            mu, sigma = r.mean(), r.std(unbiased=False)
            update = (group.G1 - mu * group.G2) / (group.size * (sigma + eps))
            batch_grad += finish_group_backward(group, update)
            group.close()

    # 4. Complete the logical batch before any parameter mutation.
    rollout.wait_batch_closed(batch)
    assert all(group.closed for group in groups.values())
    wait_for_all_backward_and_reduction_work()
    global_grad = all_reduce_sum(batch_grad)  # Sum of group directions
    write_parameter_grads(params, global_grad)
    torch.nn.utils.clip_grad_norm_(params, max_norm)
    optimizer.step()  # Exactly one ascent step at the batch barrier
    optimizer.zero_grad(set_to_none=True)

    # 5. The next rollout batch starts only after every replica ACKs.
    rollout.publish_and_wait(model, version + 1)
    return version + 1
\end{lstlisting}

\clearpage
\subsection{On-policy distillation}
\label{app:implementation-opd}

Implementation~\ref{impl:thundersync-opd} follows the turn-level OPD path. Each
teacher score uses the sampled action's realized causal context; scoring
may run upstream of the learner, but is shown inline here. The detached
student log probabilities come from the same policy snapshot, so the
importance ratio equals one while retaining its derivative. The loss
gradient is the negative of Equation~\ref{eq:opd-source-gradient}.
Turn backward captures adjoints at cut state boundaries; trajectory and
group closure propagate the remaining adjoints before the batch update.
The final sum divided by the global scored-token count is equivalent to
the runner's rescaling followed by a data-parallel average.

\begin{lstlisting}[style=thunderSyncImplementation,label={impl:thundersync-opd},caption={\sysname on-policy distillation update. Teacher-scored turns release backward immediately. Deferred context gradients and global token normalization finish before one optimizer step.}]
def thundersync_opd_step(student, teacher, rollout, batch, optimizer, max_norm):
    """One token-mean OPD update; teacher is frozen, optimizer descends."""
    params = trainable_parameters(student)
    version = rollout.policy_version
    groups = register_groups(batch)
    token_count = 0
    optimizer.zero_grad(set_to_none=True)

    # 1. Ingest turns as rollout workers generate actions and call tools.
    for event in rollout.turn_and_completion_events(batch, version):
        assert event.policy_version == version
        group = groups[event.group_id]
        if event.kind == "turn":
            with torch.no_grad():
                teacher_lp = teacher.score_action_tokens(event)
            logp = append_turn_and_get_scored_logprobs(student, event)
            if logp.numel() == 0:  # Unscored context still updates state
                continue
            old_logp = logp.detach()  # Same fixed student snapshot
            advantage = (teacher_lp - old_logp).detach()
            ratio = torch.exp(logp - old_logp)
            loss = -(ratio * advantage).sum()

            # 2. Backward now; retain adjoints needed by earlier context.
            backward_with_boundary_capture(loss, params)
            token_count += logp.numel()

        elif event.kind == "trajectory_complete":
            # 3. Replay deferred state adjoints through completed turns.
            replay_deferred_state_adjoints(event.trajectory_id)
            group.close_trajectory(event.trajectory_id)
            if group.all_trajectories_closed():
                finish_shared_prefix_backward(group)
                group.close()

    # 4. All turns, replay work, and registered groups must be complete.
    rollout.wait_batch_closed(batch)
    assert all(group.closed for group in groups.values())
    wait_for_all_backward_and_reduction_work()
    local_grad = materialize_parameter_grads(params)
    global_grad = all_reduce_sum(local_grad)
    global_tokens = all_reduce_sum(token_count)
    assert global_tokens > 0
    write_parameter_grads(params, global_grad / global_tokens)
    torch.nn.utils.clip_grad_norm_(params, max_norm)
    optimizer.step()  # Exactly one descent step at the batch barrier
    optimizer.zero_grad(set_to_none=True)

    # 5. Publish the new policy before admitting the next logical batch.
    rollout.publish_and_wait(student, version + 1)
    return version + 1
\end{lstlisting}
\endgroup

\end{document}